\documentclass{article}

\usepackage{microtype}
\usepackage{graphicx}
\usepackage{subfigure}
\usepackage{booktabs} 

\usepackage{hyperref}

\usepackage[accepted]{icmlw2019generalization}

\usepackage{amsfonts,amsmath,amssymb,bbold} 
\usepackage{array,multirow,rotating,siunitx}

\newcommand\blfootnote[1]{%
  \begingroup
  \renewcommand\thefootnote{}\footnote{#1}%
  \addtocounter{footnote}{-1}%
  \endgroup
}

\newcommand{\ba}{\mathbf{a}}

\newcommand{\bp}{\mathbf{p}}
\newcommand{\br}{\mathbf{r}}
\newcommand{\bt}{\mathbf{t}}

\newcommand{\bx}{\mathbf{x}}
\newcommand{\by}{\mathbf{y}}

\newcommand{\loss}{\ensuremath{\mathcal{L}}}
\newcommand{\error}{\ensuremath{\mathcal{E}}}
\newcommand{\cost}{\ensuremath{\mathcal{C}}}
\newcommand{\xform}{\ensuremath{\mathcal{T}}}
\newcommand{\SAP}{\ensuremath{\mathcal{P}}}

\newcommand{\spacetext}[1]{\hspace*{4mm} \text{#1} \hspace*{4mm}}

\newcommand{\header}[1]{\noindent\textbf{#1}\hspace{1em}}

\usepackage{color}

\newif\ifcomments
\commentstrue
\ifcomments
	\newcommand{\carlo}[1]{ \textcolor{red}{[#1]}}
\else
	\newcommand{\carlo}[1]{}
\fi

\ifcomments
	\newcommand{\shuzhi}[1]{ \textcolor{blue}{[#1]}}
\else
	\newcommand{\shuzhi}[1]{}
\fi

\icmltitlerunning{Identity Connections in Residual Nets Improve Noise Stability}

\begin{document}

\twocolumn[
\icmltitle{Identity Connections in Residual Nets Improve Noise Stability}




\begin{icmlauthorlist}
\icmlauthor{Shuzhi Yu}{duke}
\icmlauthor{Carlo Tomasi}{duke}
\end{icmlauthorlist}

\icmlaffiliation{duke}{Department of Computer Science, Duke University, Durham, NC, USA}

\icmlcorrespondingauthor{Shuzhi Yu}{shuzhiyu@cs.duke.edu}

\icmlkeywords{Deep Learning, Residual Neural Networks, Image Classification, Machine Learning, Noise Stability}

\vskip 0.3in
]

\blfootnote{Material in this thesis is based upon work supported by the National Science Foundation under Grant No. CCF-1513816.}



\printAffiliationsAndNotice{}  

\begin{abstract}
Residual Neural Networks (ResNets) achieve state-of-the-art performance in many computer vision problems. Compared to plain networks without residual connections (PlnNets), ResNets train faster, generalize better, and suffer less from the so-called degradation problem. We introduce simplified (but still nonlinear) versions of ResNets and PlnNets for which these discrepancies still hold, although to a lesser degree.  We establish a 1-1 mapping between simplified ResNets and simplified PlnNets, and show that they are exactly equivalent to each other in expressive power for the same computational complexity. We conjecture that ResNets generalize better because they have better noise stability, and empirically support it for both simplified and fully-fledged networks.
\end{abstract}

\section{Introduction}
Deep convolutional neural networks have rapidly improved their performance in image classification tasks in recent years. So-called \emph{residual networks}~\cite{He_2016_CVPR}, in particular, differ from regular ones by the mere addition of an identity function to certain layers. It has been shown this small change makes residual networks train faster than their plain counterparts, in that they converge to the same training risk in fewer epochs, and  reduce the so-called \emph{degradation problem}~\cite{He_2016_CVPR}, that is, their training risk increases more slowly as new layers are added. In addition, ResNets generalize better, that is, they achieve lower classification risk for the same training risk.


In an attempt to understand why a small change in network architecture leads to the advantages listed above, we introduce a formal transformation between simplified versions of residual networks and their nearest kin, which we call (simplified) \emph{plain networks}. Specifically, for every simplified residual network we show how to construct a simplified plain network of equal parametric complexity, and \textit{vice versa}, such that the two networks are exactly equivalent to each other, if exact arithmetic is used. By ``equivalent'' we mean that the two networks implement the same input/output relation.
Thus, the differences in performance between plain and residual networks relate to the initialization and training, instead of expressive power. In summary, \emph{simplified residual networks are equivalent to simplified plain networks of equal parametric complexity initialized in different ways for training.}

The simplification we made relative to the networks used in the literature amounts to removing batch normalization layers and allowing for skip connections (the identity added to residual networks) only across individual layers, rather than across groups of layers. However, our simplified architectures exhibit qualitatively similar differences between plain and residual versions as the fully-fledged ones do, albeit to a lesser degree. Because of this, our insights may help understand commonly used architectures also, a claim we support with experiments on fully-fledged networks.

To explain the advantages above, we observe that adding an identity across a layer makes some of the layer's weights close to 1, a value much higher than the small random values used for initializing either type of network (plain or residual). We show empirically that the large weights present in residual networks decrease the relative sensitivity of the output of a network to changes in layer parameters. It has been shown recently~\cite{Ge2018ICML} that such a decrease in sensitivity (increase in \emph{noise stability}) is a good predictor of improved generalization. In addition, we empirically observe better noise stability in the fully-fledged residual networks as well. 


Some recent theoretical work analyzes linear ResNets, from which ReLUs are removed~\cite{Moritz_Tengyu,kawaguchi2016deep, li2016demystifying, zaeemzadeh2018norm} or analyzes shallow nonlinear ResNets \cite{Orhan2018iclr}. In addition, alternative views of ResNets have been proposed, such as the \emph{unraveled view} \cite{veit16residual, littwin2016loss, huang2016deep, abdi2016multi}, the \emph{unrolled iterative estimation view} \cite{greff2017iclr, jastrzebski2018iclr}, the \emph{dynamic system view} \cite{chang2018multilevel, Chang2018ReversibleAF, liao2016bridging}, and the \emph{ensemble view} \cite{icmlhuang18}. In contrast, we analyze simplified but deep and non-linear networks. Our work does not reason by analogy, but instead explores properties of skip connections through a mapping between ResNets and PlnNets.

Previous work has shown the importance of weight initialization for training deep neural networks and proposed good initialization algorithms \cite{Mishkin2016LSUV, he2015delving, GlorotAISTATS2010, philipp2016}. We show that a ResNet is a PlnNet with a special initialization, and that the resulting weights, which involve entries much larger than in previous studies, improve generalization ability. 

\header{Noise stability} A recent paper~\cite{Ge2018ICML} introduces bounds for the ability of a network to generalize that are tighter than standard measures of hypothesis-space complexity are able to provide. They compress a network to a simpler, approximately equivalent one, and tie generalization bounds to a count of the number of parameters of the simpler network. Compressibility, and therefore generalization, improve if the network is \emph{noise-stable}, in the sense that perturbations of its parameters do not change the input/output function implemented by the network too much. We show empirically that (fully fledged) ResNets are more noise-stable than their plain counterparts.

\section{Simplified PlnNets and ResNets}
\label{pr_nets}
A \emph{Plain Neural Network} (PlnNet) and a \emph{Residual Neural Network} (ResNet) are the concatenation of \emph{plain blocks} and \emph{residual blocks} respectively, each of which computes a transformation between input $\bx$ and output $\by$ of the following form respectively:
\[
	\by = f(p(\bx, \bp))\;\;\; \mbox{and}\;\;\;
	\by = f(\bx + p(\bx, \br))
\]
where function $f$ is a ReLU nonlinearity. In this expression, $p$ is a cascade of one or more of the standard blocks used in convolutional neural networks, and has parameter vector $\bp$ and $\br$ respectively. Thus, a residual block merely adds an identity connection from the input $\bx$ to the output $p$ of a plain block.



A specific architecture of the residual block that contains two convolutional layers has been shown empirically to outperform several other variants. However, the goal of our investigation is not to explain the performance of the best architecture, but rather to compare plain networks with residual networks. We have found empirically that even a drastically simplified block architecture (shown in Figure~\ref{fig:pWOB_rWOB_block}) preserves, to a lesser but still clear extent, the advantages of ResNets over PlnNets mentioned above (see Supplementary Material (SM) Section~\ref{baseline_exp}). For example, the classification error for a 20- and 44-layer simplified PlnNets are 0.1263 and 0.2213 respectively. In comparison, those of a 20- and 44-layer simplified ResNets are 0.1131 and 0.1182 respectively. Because of this, in this paper we compare cascades of \emph{depth-1} plain blocks of the form in Figure \ref{fig:pWOB_rWOB_block}(a) with cascades of depth-1 residual blocks of the form in Figure \ref{fig:pWOB_rWOB_block}(b).

More specifically, PlnNets (or ResNets) of $L$ layers are for us the concatenation of a convolutional layer with ReLU, $3N$ plain (or residual) blocks, an average pooling layer and a final, fully-connected layer followed by a soft-max layer (more details in SM Figure \ref{fig:pr_arch}). All the filters have size $3\times 3$ in our architecture.

\begin{figure}[!h]
    \vskip -0.15in
    \centering
    \subfigure[]{
      \includegraphics[width=0.35\columnwidth]{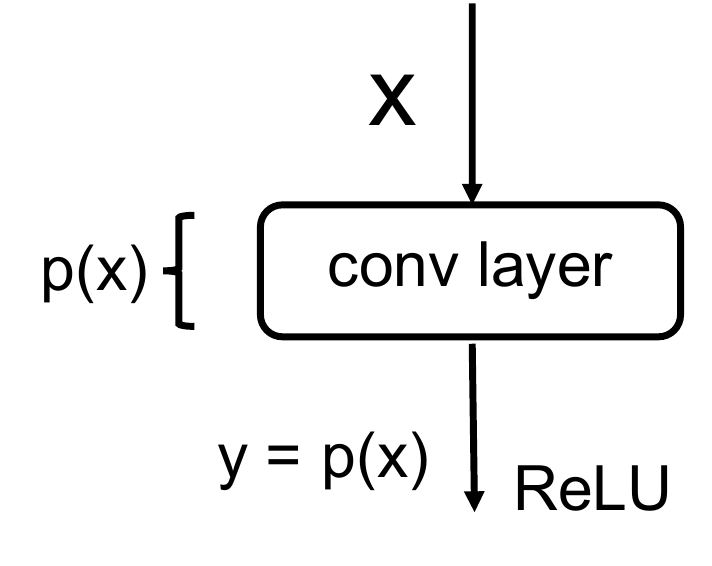}}
      \quad \quad
    \subfigure[]{
      \includegraphics[width=0.45\columnwidth]{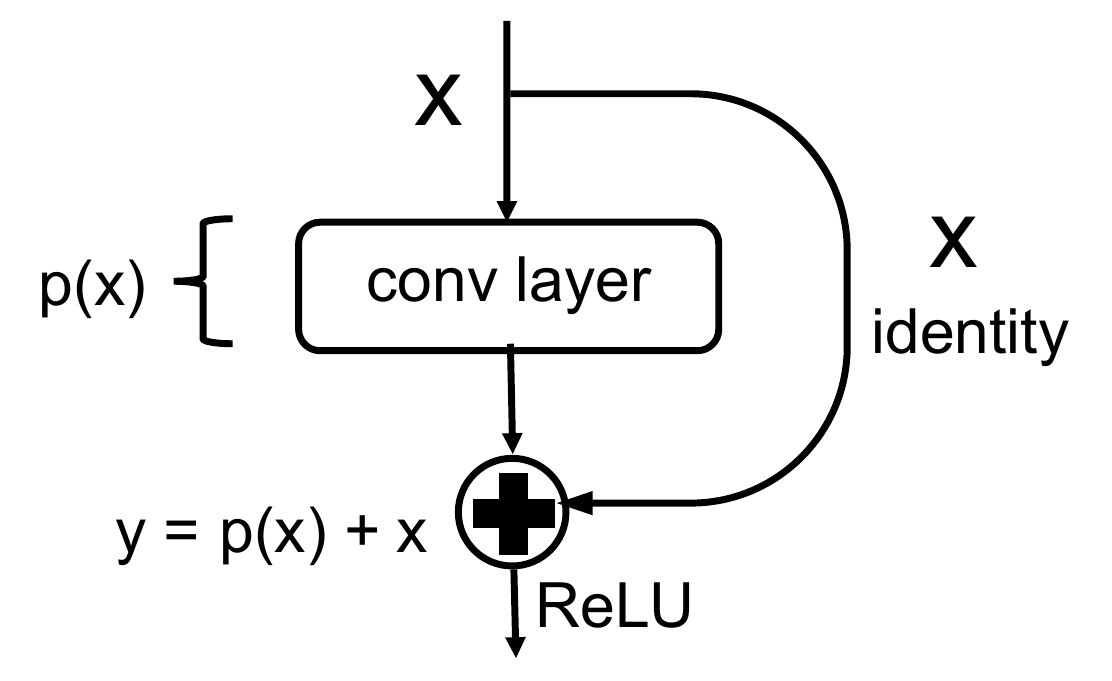}}
  \caption {Simplified (a) plain block and (b) corresponding residual block.}
  \label{fig:pWOB_rWOB_block}
  \vskip -0.15in
\end{figure}

\section{Simplified PlnNets and ResNets are Equivalent}
\label{eq_sim_ps_nets}
We now show that \emph{the simplified plain and residual blocks are exactly equivalent to each other}, in the sense that a plain block with convolutional layer $p(\bx, \bp)$ can be made to have the exact same input/output relationship of a residual block with convolutional layer $r(\bx, \br) = \bx + p(\bx, \br)$ as long as the parameter vectors $\bp$ and $\br$ relate to each other as follows.

Consider a plain network block $p$ with $C$ input channels, whose convolutional layer has $D$ kernels (and therefore $D$ output channels) of width $U = 2U_h + 1$ and height $V = 2V_h + 1$ for nonnegative integers $U_h$ and $V_h$ (the \emph{half-widths} of the kernel). Then, the set of kernels can be stored in a $D\times C \times U \times V$ array. We index the first two dimensions by integers in the intervals $[1, D]$ and $[1, C]$, and the last two in the intervals $[-U_h, U_h]$ and $[-V_h, V_h]$.

We define the \emph{ID entries} of that set of kernels to be the kernel coefficients in positions $(c, c, 0, 0)$ of the array, for $c = 1,\ldots, \min(C, D)$. These are the entries that identities are added to in residual networks, and therefore the plain network $p$ and residual network $r$ implement the same function if
\begin{equation}
\bp \;=\; \xform(\br) \;=\; \br + {\boldsymbol\rho} \;\;\;\mbox{where}\;\;\;
{\boldsymbol\rho} = \left\{\begin{array}{ll}1 & \mbox{for all ID entries}\\
0 & \mbox{elsewhere.} \end{array}\right.
\label{eq:modDecay}
\end{equation}
Of course this transformation is easily invertible:
\[
\br \;=\; \xform^{-1}(\bp) \;=\; \bp - {\boldsymbol\rho}
\]
so that a plain network can be transformed to an equivalent residual network as well.

The transformations $\xform$ and $\xform^{-1}$ are extended from blocks to networks by applying the block transformation to each of the network's $3N$ blocks in turn.

\emph{A crucial consequence of this equivalence is that the better performance of residual networks over plain networks, which manifests itself even when the networks have their simplified form, does not derive from differences in expressive power between the two architecture. 
\textbf{Simplified residual networks are simplified plain networks that are trained into a different local minimum of the same optimization landscape.}}

\section{Training Plain and Residual Networks}
\label{train_pr_nets}

As commonly done, we define the loss of a network as the cross-entropy between prediction and truth at the output of the final soft-max layer, and the training risk is the average loss over the training set. In practice, it is common in the literature to add \emph{weight decay}, an $L_2$ regularization term, to the risk function to penalize large parameter values and improve generalization.~\cite{hinton1987,krogh1992} 

\subsection{Equivalent Training of Equivalent Networks}

With slight abuse of notation, we henceforth denote with $p$ and $r$ the transformation performed by an entire network (plain or residual), rather than by a single block.

If a plain network $p$ and the corresponding residual network $r$ have parameter vectors $\hat{\bp}$ and $\hat{\br}$, respectively, the functions $p(x, \hat{\bp})$ and $r(x, \hat{\br})$ the two networks implement are different. However, if the initial parameter vectors $\bp_0$ and $\br_0$ satisfy the equation
\begin{equation}
\bp_0 = \xform(\br_0)\;,
\label{eq:initEquiv}
\end{equation}
the two networks $p(\bx, \bp_0)$ and $r(\bx, \br_0)$ are input-output-equivalent to each other.

In addition, since subsets of corresponding layers in $p$ and $r$ are also equivalent to each other, inspection of the forward and backward passes of back-propagation shows immediately that the gradient of the training risk function with respect to $\bp$ in $p$ is the same as that with respect to $\br$ in $r$. Thus, if training minimizes the training risk, and the sequence of training sample mini-batches is the same for both training histories, a plain network $p$ and a residual network $r$ whose initial weights satisfy equation \ref{eq:initEquiv} remain equivalent at all times throughout training.

To ensure equivalence between plain and residual networks even in the presence of weight decay, the penalty $\|\bp\|^2$ for residual networks is replaced by $\|\xform^{-1}(\bp)\|^2$. In this way, if plain network $p$ and residual network $r$ are equivalent, both their costs and the gradients of their cost are equal to each other as well (experiment details in SM Section \ref{eq_train}).
We can therefore draw the following conclusion.

\emph{If exact arithmetic is used, training plain network $p$ with initialization $\bp_0 = \xform(\br_0)$ and weight-decay penalty $\|\xform^{-1}(\bp)\|^2$ is equivalent to training residual network $r$ with initialization $\br_0$ and weight-decay penalty $\|\br\|^2$. If the two networks are trained with the same sequence of mini-batches, the plain and residual networks $p(\bx, \bp_e)$ and $r(x, \br_e)$ at epoch $e$ during training are equivalent to each other for every $e$, and so are, therefore, the two networks obtained at convergence.}

\subsection{ResNets are PlnNets with Large ID Entries}
The equivalence established in the previous Section allows viewing a simplified residual network as a simplified plain network with different weights:
\[
r(\bx, \br) = p(\bx, \xform(\br))\;.
\]
Because of this, instead of comparing a plain network $p(\bx, \bp)$ with a residual network $r(\bx, \br)$, we can compare the two plain networks $p(\bx, \bp)$ and $p(\bx, \xform(\br))$ to each other. These have the exact same architecture and landscape, but their parameter vectors are initialized with different weights, and end up converging to different local minima of the same risk function.

A first insight into this comparison can be gleaned by comparing the distributions of network weights (excluding biases) for $p(\bx, \bp_e)$ and $p(\bx, \xform(\br_e))$ at epoch $e$ of training. For notational simplicity, we define the \emph{transferred parameter vector} $\bt = \xform(\br)$. We initialize the two networks $\bp_0$ and $\bt_0$ with Kaiming Weight Initialization (KWI) method~\cite{he2015delving} and Hartz-Ma Weight Initialization (HMWI) method~\cite{Moritz_Tengyu} respectively (Details in SM Section \ref{baseline_exp}). Since network equivalence is preserved during training, we can also write $\bt_e = \xform(\br_e)$ for every epoch $e$, so we compare plain networks $p(\bx, \bp_e)$ and $p(\bx, \bt_e)$.

Initially ($e = 0$), the distribution $\chi(z, \bp_0)$ of $\bp_0$ is a zero-mean Gaussian distribution with variance $\sigma_p^2$, for the weights. The distribution $\chi(z, \bt_0)$ of the initial transferred weights $\bt_0$ is a mixture of two components: A Gaussian with zero mean and variance $\sigma_r^2$ for weights other than the ID entries, and a Gaussian with mean 1 and variance $\sigma_r^2$ for the ID entries. These distributions change during training. Incidentally, and interestingly, we have observed empirically that these changes are small relative to 1 for the example of networks $p(\bx, \bp_e)$ and $p(\bx, \bt_e)$ of depth 32, so training does not move weights very much within the optimization landscape (more details in SM Section \ref{dist_ev}). 

Stated differently: 
\emph{The training histories for (i) a plain network initialized by KWI and (ii) a plain network of equal architecture but equivalent to a residual network initialized by HMWI evolve in relatively small neighborhoods of their starting points in parameter space. The two neighborhoods are far apart, separated in large measure by the size of the ID entries in the two networks.}

\section{Large ID Entries Improve Noise Stability}
\label{noise_stability}
The noise stability of a network~ \cite{Ge2018ICML} is measured by how much the output changes when noise is added to the weights of some layer. More specifically, the \emph{cushion} for layer $\ell$ is an increasing function of a layer's noise stability, and is defined as the largest number $\mu_{\ell}$ such that for any data point $\bx$ in the training set,
\[
\mu_{\ell}\left\|A^{\ell}\right\|_F \left\|\bx^{\ell-1}\right\| \leq \left\|A^{\ell} \bx^{\ell-1}\right\|\;.
\]
In this expression, $A^{\ell}$ represents the weights in the layer, $\bx^{\ell-1}$ is the output of previous layer (after the ReLU), and $\|\cdot\|$ and $\|\cdot\|_F$ are the 2-norm and Frobenius norm.

Because plain networks equivalent to residual networks contain large weights, the same amount of noise added to such a network has a smaller effect on the output than for a plain network initialized in the traditional way. Thus, residual networks have greater layer cushions and greater noise stability overall, which leads to better generalization when achieving the same loss. As shown in Figure~\ref{fig:pspt20_losst}, with same training loss, the ResNets achieve smaller error.

\begin{figure}[h]
    \centering
    \subfigure[]{\includegraphics[width=0.45\columnwidth]{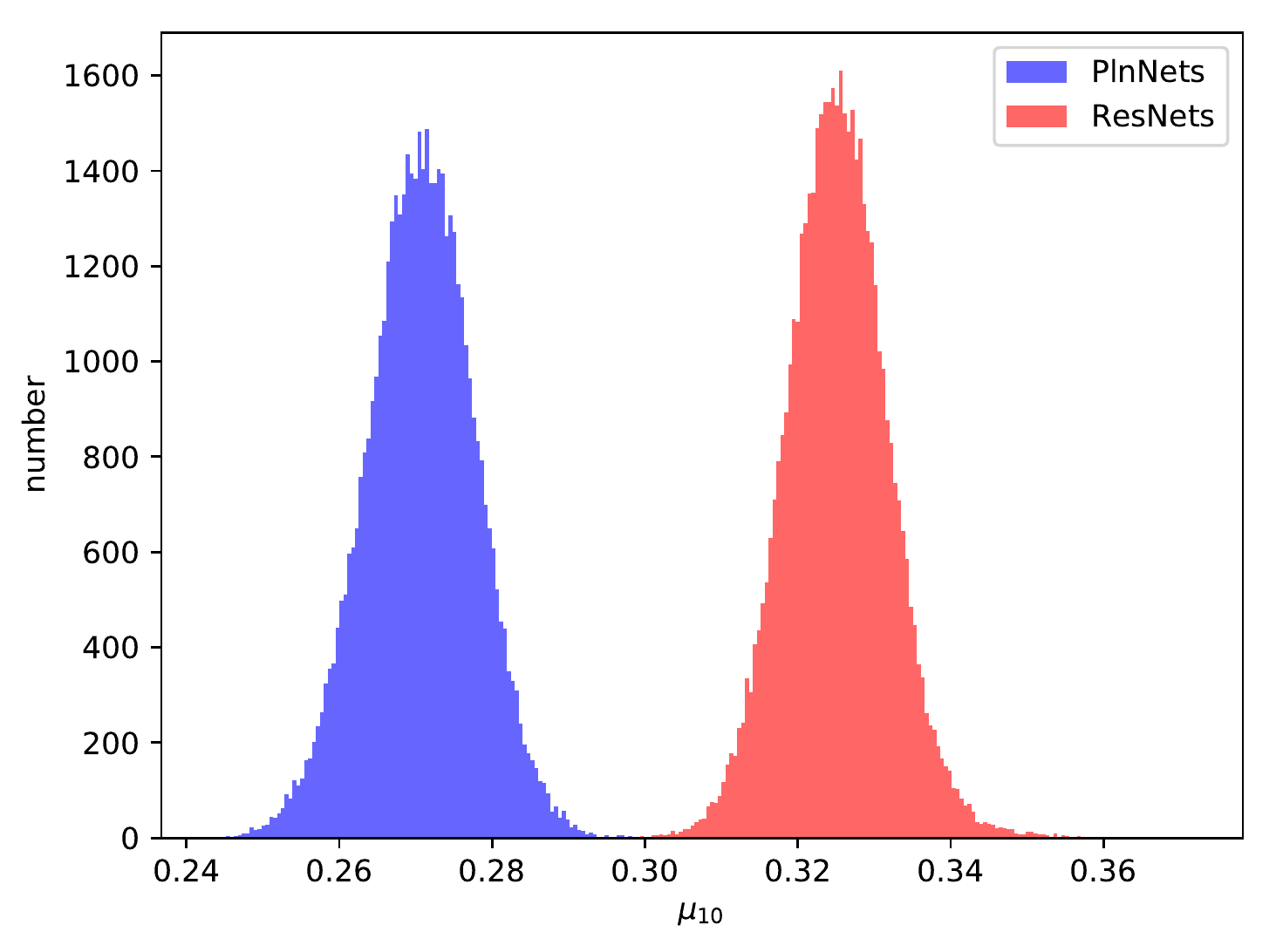}}
    \subfigure[]{\includegraphics[width=0.45\columnwidth]{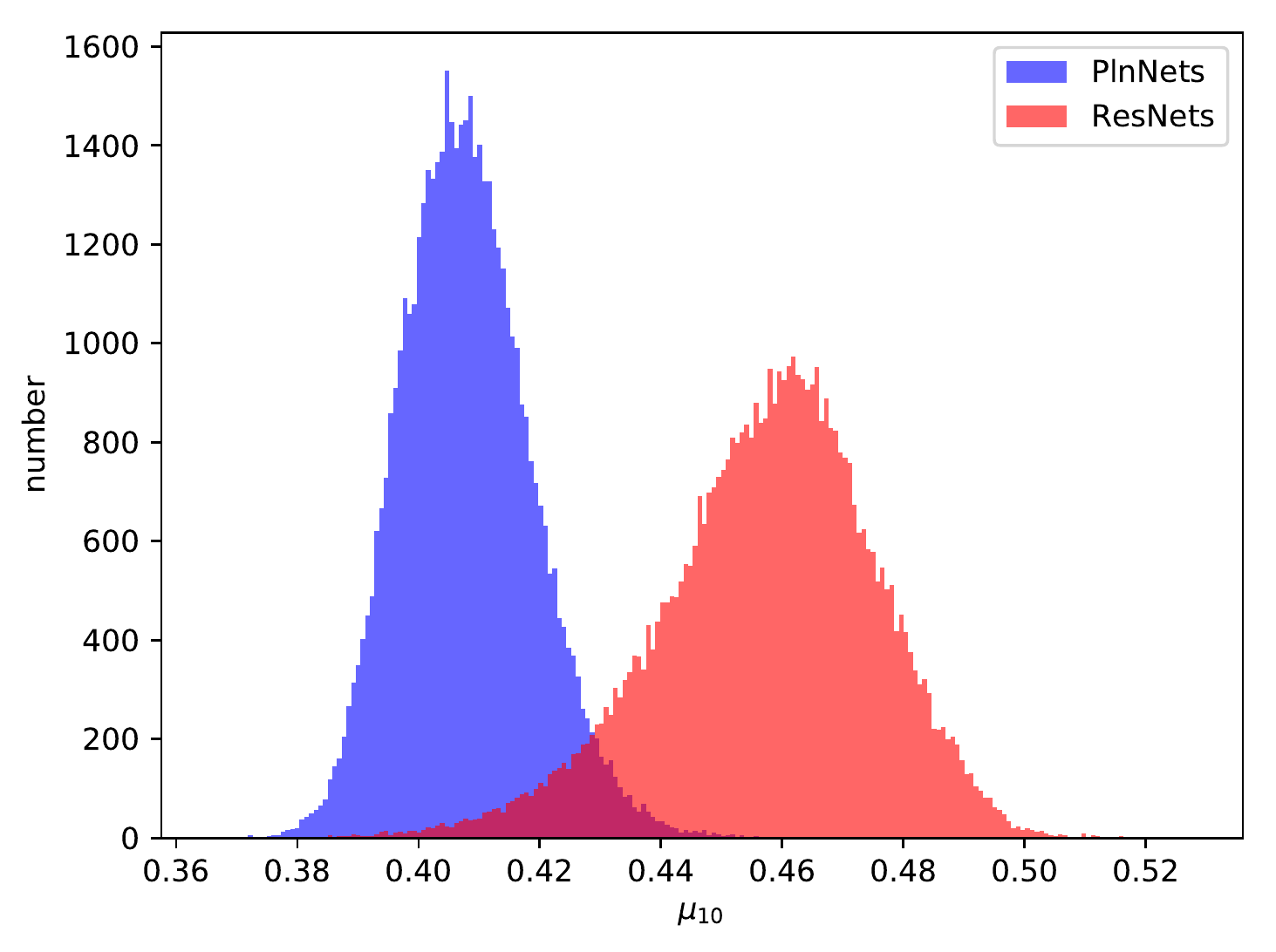}}
    \caption {Histogram of the per-sample layer cushion in the $10^{th}$ layer of 20-layer (left) and 32-layer (right) PlnNets (blue) and ResNets (red). All the four networks are trained for 200 epochs on CIFAR-10 dataset. The test errors for 20-layer PlnNet and ResNet are 0.1263 and 0.1122 respectively, and those for 32-layer networks are 0.1349 and 0.1121 respectively.}
\label{fig:layer_cushion}
\vskip -0.15in
\end{figure}

Empirical measurements support this view. Specifically, Figure~\ref{fig:layer_cushion} shows the distribution of the per-sample cushion $\mu_{\ell}(\bx^{\ell-1}) = \left\|A^{\ell} \bx^{\ell-1}\right\|/\left\|A^{\ell}\right\|_F /\left\|\bx^{\ell-1}\right\|$ for the $10^{th}$ layer of a pre-trained 20-layer and 32-layer plain and residual network. To make this Figure, we measure the Frobenius norm of the weights of ResNets after removing 1 from the ID entries, because the intent of the term $\left\|A^{\ell}\right\|_F$ is to measure \emph{change} in the coefficients. The plots in the figure show that ResNets have better noise stability than PlnNets do, as the red distribution is to the right of the blue distribution. This pattern holds for all the layers, with the exception of 2 layers (in both the 20-layer and 32-layer networks). This result implies~\cite{Ge2018ICML} that ResNets are more likely to generalize better than PlnNets do. In summary, the large ID entries make ResNets less vulnerable to noise in its weights, leading to better generalization.

The better noise stability holds not only for simplified networks, but also for the fully-fledged ones. To show this, we measure the per-sample interlayer cushion $\mu_{i,j}(\bx^i)$ of plain and residual blocks in commonly used networks. Given layers $i \leq j$,
\[
\mu_{i,j}(\bx^i) = \sqrt{n^{(i)}} \left\|J^{i, j}_{\bx^i} \bx^i\right\| / \left\|J^{i, j}_{\bx^i}\right\|_F / \left\|\bx^i\right\|
\]
where $n^{(i)}$ is the size of the input $\bx^i$ to layer $i$, and $J^{i, j}_{\bx^i}$ is the Jacobian of the part of the network from layer $i$ to layer $j$. Experimental results show that $\mu_{i,j}(\bx^i)$ for a residual block is much bigger than that of the corresponding plain block. For example, the interlayer cushion of the $5^{th}$ out of 9 residual block of a fully-fledged residual network of depth 20 has a mean 0.20 and standard deviation 0.0024. In comparison, the interlayer cushion of the corresponding plain block has a mean $2.7\times 10^{-7}$ and standard deviation $2.4\times 10^{-8}$. The plain block in this case is the concatenation of two groups of convolutional layer, Batch Normalization layer, and ReLU; the residual block adds the input of the first convolutional layer to the output of second Batch Normalization layer of the plain block. 

\begin{figure}[h]
	\vskip -0.15in
    \centering
    \subfigure[]{\includegraphics[width=0.5\columnwidth]{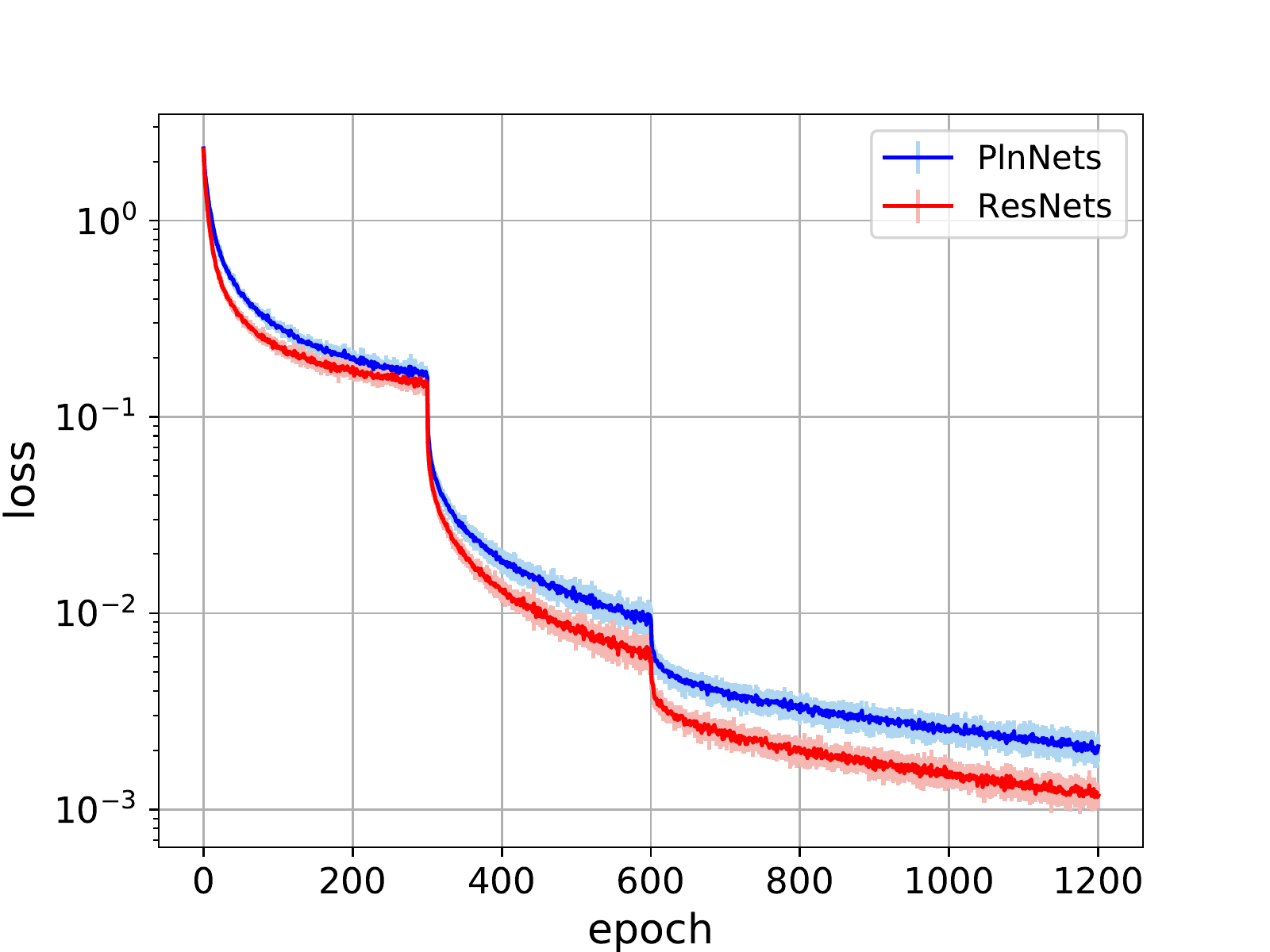}}
    \subfigure[]{\includegraphics[width=0.48\columnwidth]{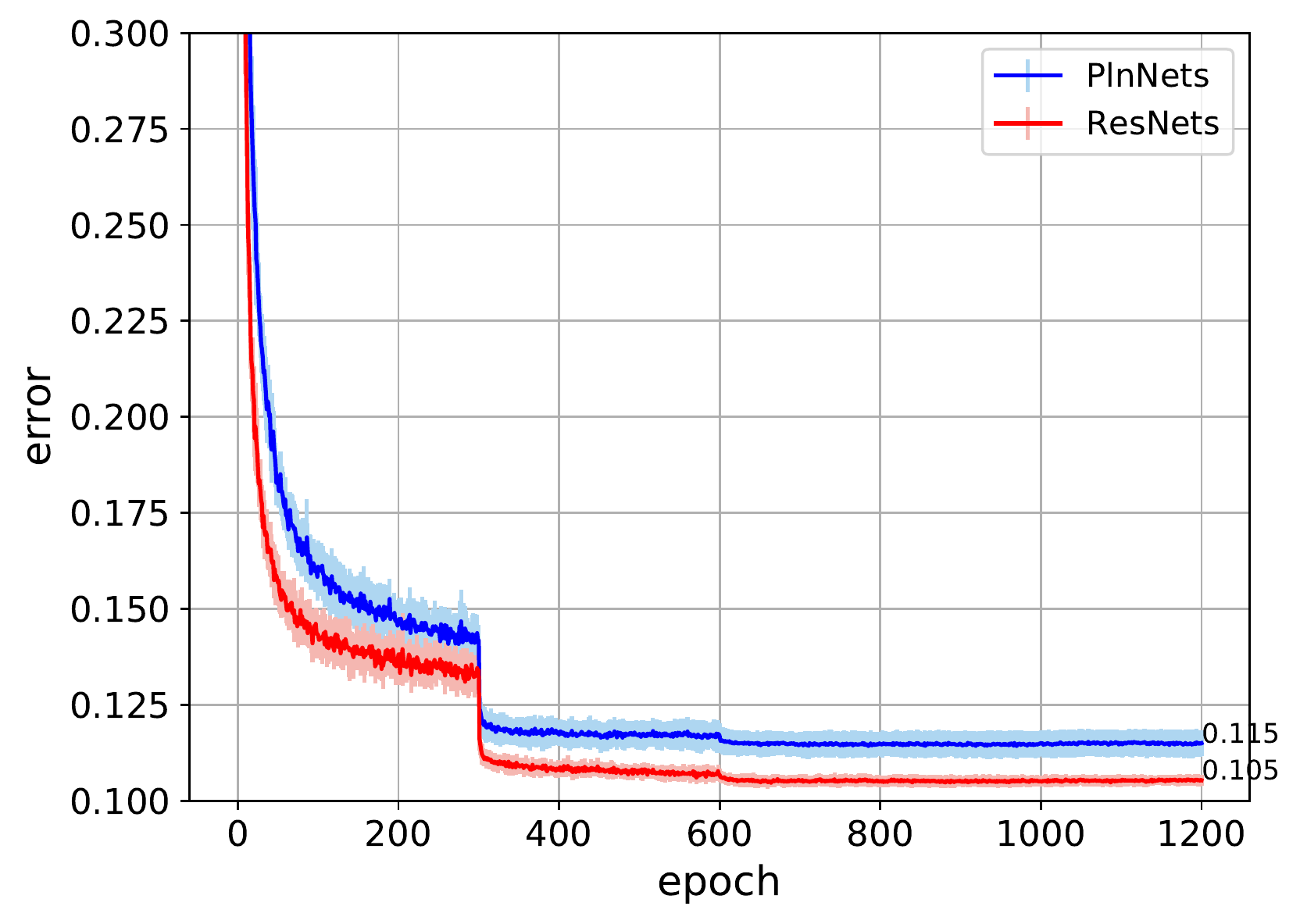}}
    \caption {Plots of risk $\loss$ (left) and error $\error$ (right) as a function of training time for plain (blue) and residual (red) networks of depth 20 after training for 1200 epochs on the CIFAR-10 dataset. The results are based on training 13 plain networks and 13 residual networks initialized at random (but equivalently to each other), to estimate variance in the results. The sequence of training samples are also randomized. Dark color refers to the mean and the half length of the error bar represents the standard deviation.
}
\label{fig:pspt20_losst}
\vskip -0.15in
\end{figure}

\section{Conclusions}
\label{conclusions}
We propose a one-to-one map between simplified ResNets and PlnNets, and we show how to train corresponding networks equivalently. We conjecture that ResNets achieve lower generalization error than PlnNets because the large ID entries make the former more stable against noise added to the weights. Our experiments support our conjectures for both simplified and fully-fledged networks.

\nocite{langley00}


\appendix

\section{Empirical Baseline}
\label{baseline_exp}
\begin{figure}[h]
    \centering
    \subfigure[]{\includegraphics[width=0.38\columnwidth]{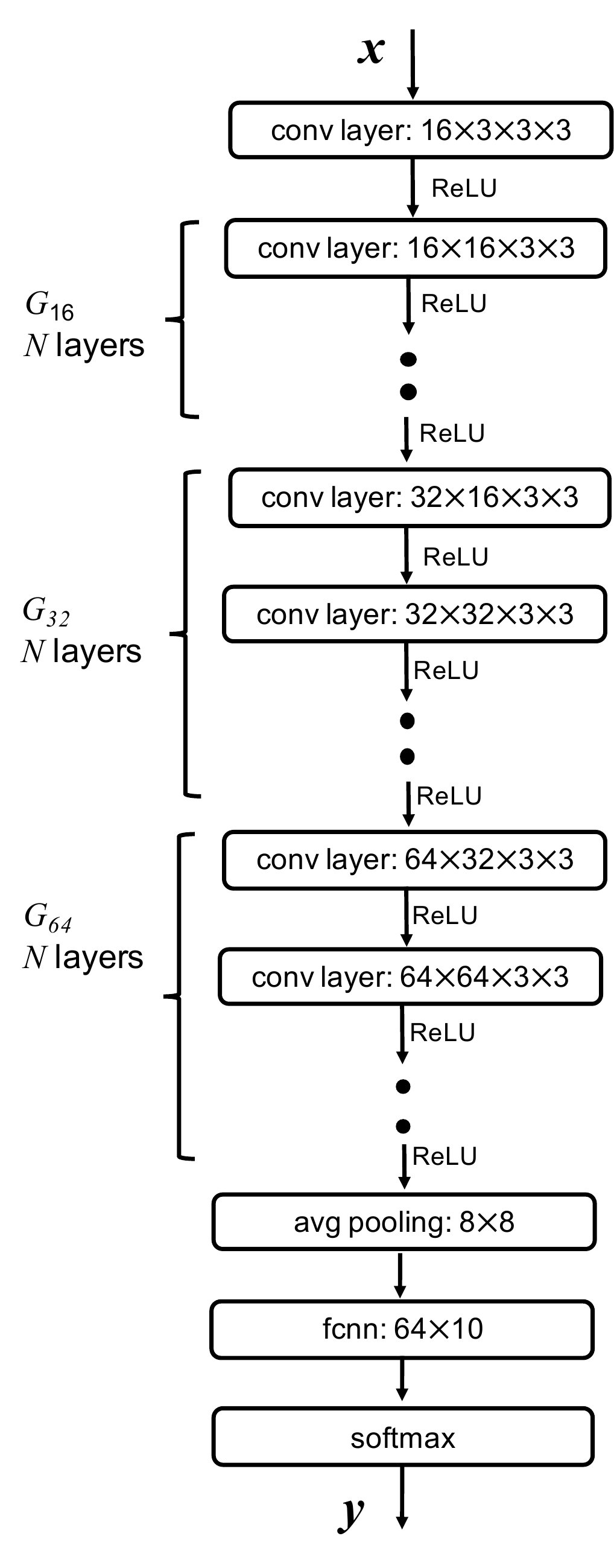}}
    \quad \quad
    \subfigure[]{\includegraphics[width=0.36\columnwidth]{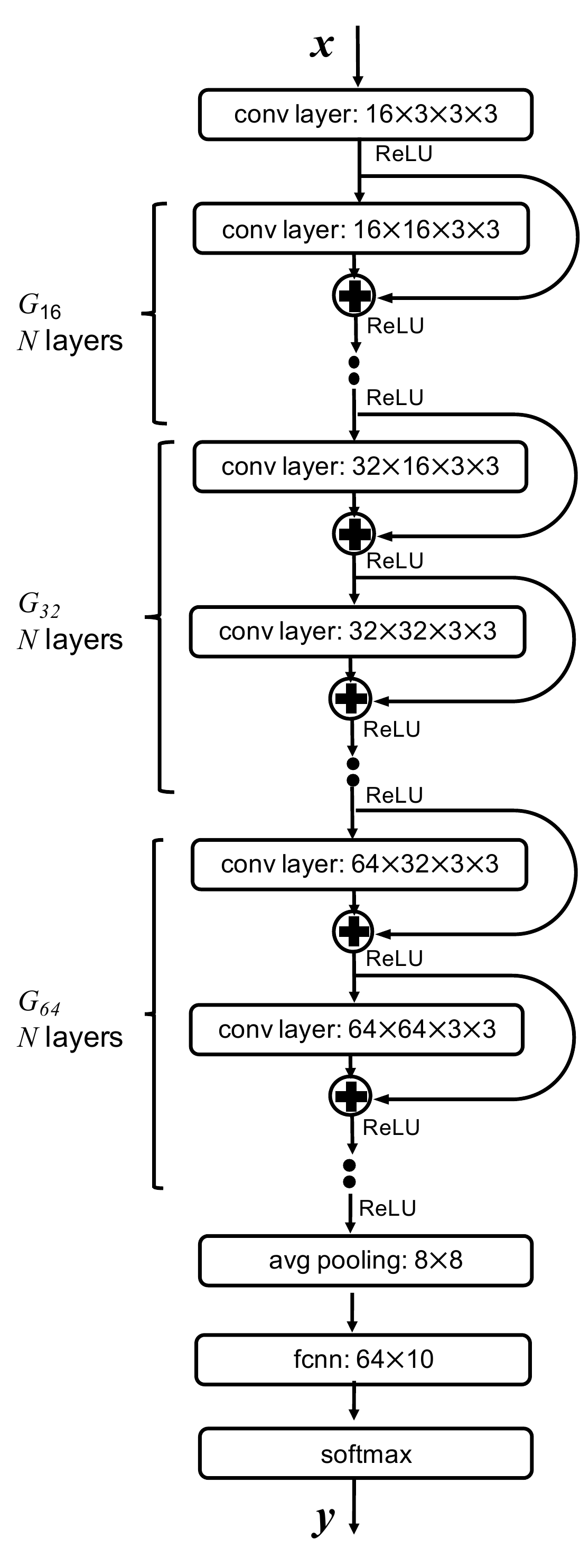}}
  \caption {The architecture of (a) plain network and (b) the corresponding residual network. There are three groups of $N$ layers each: $G_{16}$, $G_{32}$, and $G_{64}$. The entire network has $L = 3N+2$ layers. The number of weights of both networks of same depth equals. The inputs from CIFAR-10 data set are $32\times 32 \times 3$ RGB images and the output is a vector of size $10\times 1$ that represents soft-max scores for the 10 classes.}
  \label{fig:pr_arch}
  \vskip -0.2in
\end{figure}

Our experiments are conducted on CIFAR-10 data set that contains 50k training samples and 10k testing samples from 10 classes. Each image in the data set is RGB image of size $32\times 32$. The simplified network architectures are shown in Figure~\ref{fig:pr_arch}. PlnNets (or ResNets) of $L$ layers are for us the concatenation of a convolutional layer with ReLU, three groups, namely $G_{0}$, $G_{1}$, and $G_{2}$, with $N$ plain (or residual) blocks each, an average pooling layer and a final, fully-connected layer followed by a soft-max layer. There are $L = 3N + 2$ trainable layers in total. All the filters have size $3\times 3$ in our architecture. Each of the first $N + 1$ convolutional layers has 16 filters. The first convolutional layer in both $G_{1}$ and $G_{2}$ doubles the number of output channels but halves the input width and height; the rest of the convolutional layers keep the size of input and output the same. The output of the last convolutional layer has 64 channels with each channel of size $8\times 8$. The average pooling layer has window size of $8\times 8$ so that the result after pooling layer is a vector of 64 entries. The final output of the networks is a vector of size $10\times 1$ that represents soft-max scores for the 10 classes.

As an initial empirical baseline, we trained simplified PlnNets and ResNets (whose building blocks are shown in Figure~\ref{fig:pWOB_rWOB_block} (a) and (b) respectively) of depths 20, 32, and 44. Plain networks are typically initialized by the Kaiming Weight Initialization (KWI) method~\cite{he2015delving}. In this method, biases are set to zero, and the weights of a $C\times U\times V$ convolutional kernel are set to samples from a Gaussian distribution with mean 0 and standard deviation
\begin{equation}
\sigma_p = \sqrt{\frac{2}{S}}\;\;\;\mbox{where}\;\;\; S = CUV\;.
\label{eq:sigmaP}
\end{equation}

Initialization of the parameters of a residual network is similar, except that $\sigma_p$ is replaced by
\begin{equation}
\sigma_r = \frac{1}{S}\;,
\label{eq:sigmaR}
\end{equation}
which is typically much smaller than $\sigma_p$. This prescription, called Hartz-Ma Weight Initialization (HMWI) method~\cite{Moritz_Tengyu}, reflects the observation that residual weights are generally smaller than plain weights. Table~\ref{tab:err_loss_pr203244} shows training loss $\loss$ and test error $\error$ after 200 epochs of training, when the loss stabilizes.

\begin{figure}[h]
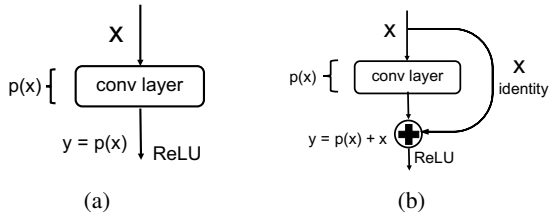

    \centering
    \subfigure[]{
      \includegraphics[width=0.35\columnwidth]{plnnetWOB_block.pdf}}
      \quad \quad
    \subfigure[]{
      \includegraphics[width=0.45\columnwidth]{resdenWOB_block.pdf}}
  \caption {Simplified (a) plain block and (b) corresponding residual block.}
  \label{fig:pWOB_rWOB_block}
\end{figure}

We observe that 1) ResNets have smaller $\loss$ and $\error$ than the PlnNets of equal depth, and the performance discrepancy between ResNets and PlnNets increases with the depth of the network. 2) PlnNets exhibit an obvious degradation problem, as $\loss$ increases 6.85 (0.4950/0.07227) times from depth 20 to depth 44. This problem is much smaller for ResNets, as $\loss$ increases 1.73 (0.07397/0.04287) times from depths 20 to 44. 3) The test error $\error$ for PlnNets increases significantly with depth, but remains roughly constant for ResNets.

In summary, our simplified ResNets outperform our simplified PlnNets in terms of both training loss and test error, and ResNets suffer much less from the degradation problem. Thus, \emph{our network simplification preserves the phenomena we wish to study.}

\begin{table*}[h!]
\centering
\begin{tabular}{|c|c|c|c|c|c|c|}
    \cline{2-7}
    \multicolumn{1}{c}{} & \multicolumn{3}{|c|}{PlnNet} & \multicolumn{3}{c|}{ResNet}\\
    \hline
    depth & $\loss$ & $\cost$ & $\error$ & $\loss$ & $\cost$ & $\error$ \\
    \hline
    20 & 0.07227 & 0.07304 & 0.1263 & 0.04287 & 0.04338 & 0.1131 \\
    \hline
    32 & 0.1225 & 0.1234 & 0.1349 & 0.04847 & 0.04903 & 0.1143 \\
    \hline
    44 & 0.4950 & 0.4962 & 0.2213 & 0.07397 & 0.07459 & 0.1182 \\
    \hline
\end{tabular}
\caption{Training loss $\loss$, training cost $\cost$ and test error $\error$ of PlnNets and ResNets for increasing depth and after training for 200 epochs on the CIFAR-10 dataset. The initial learning rate was $10^{-3}$ for the depth-44 plain network, and $10^{-2}$ for the other 5 networks. The reason for the exception is that the training loss for the depth-44 plain network does not decrease with larger learning rates. The learning rate was divided by 10 after epochs 120 and 160 for all networks. The parameter $\lambda$ of weight decay is $10^{-4}$. Loss $\loss$ is generally over 80 times more than the weight decay term. Figures are reported with four significant decimal digits.}
\label{tab:err_loss_pr203244}
\end{table*}

\section{Equivalent Training of Equivalent Nets}
\label{eq_train}
PlnNets and ResNets can only be trained equivalently in ideal case. Empirically, we observed some small discrepancies as a result of rounding errors. Specifically, Table~\ref{tab:err_loss_ptrs203244} shows the results of the same experiment performed for Table \ref{tab:err_loss_pr203244}, except that the two networks are initialized to be equivalent. We used double precision in the experiments, and performed all calculations on the same CPU to avoid possible hardware differences between different GPUs or CPUs.

\begin{figure}
    \centering
    \subfigure[]{\includegraphics[width=0.5\columnwidth]{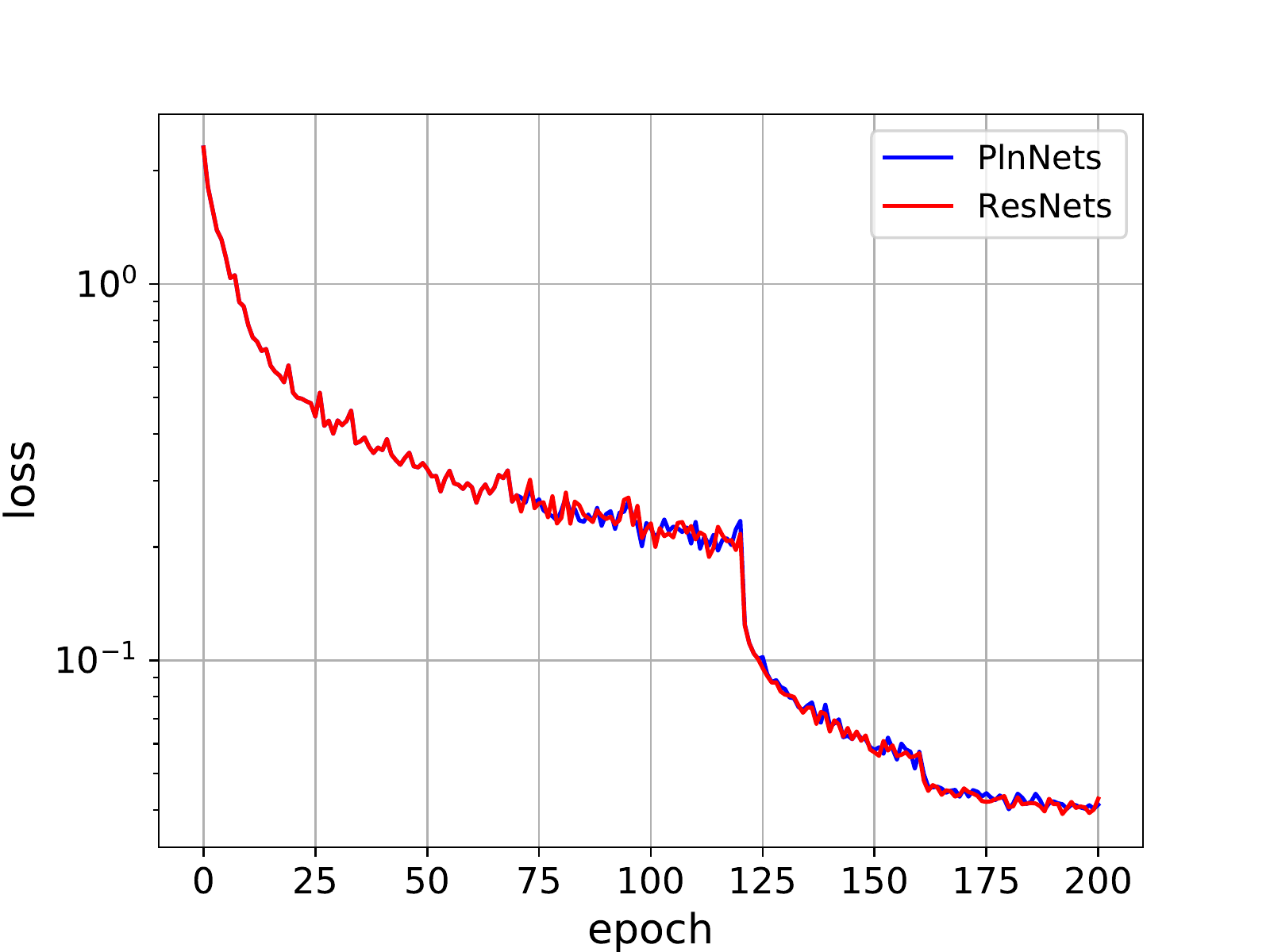}}
    \subfigure[]{\includegraphics[width=0.46\columnwidth]{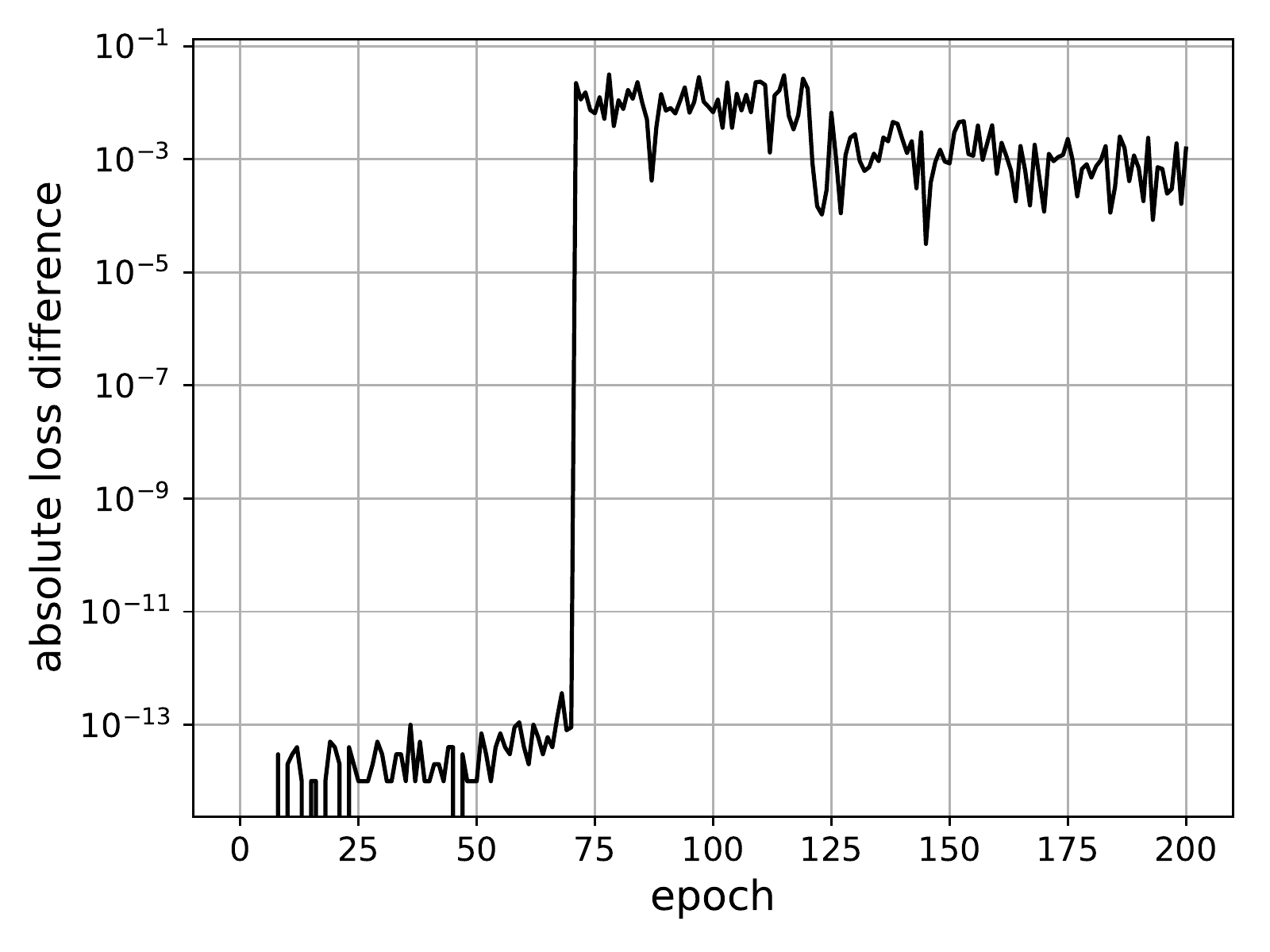}}
  \caption {Left: Semi-logarithmic plots of $\loss$ for plain (blue) and residual (red) networks of depths 20, initialized with equivalent parameter values. Right: Semi-logarithmic plots of the absolute differences.}
  \label{fig:ptrs_losst_2044}
\end{figure}

We explain these discrepancies based on the following observations. After one iteration of training, some outputs of the second convolutional layer differ by about $10^{-16}$ between the two networks, as a result of rounding errors (the relative accuracy of double-precision floating-point arithmetic is $10^{-17}$) and of the different orders of magnitude of the ID weights between plain and residual network (ID weights are of the order of $10^{-3}$ for residual networks, and the corresponding plain ID weights are close to 1). As numerical errors propagate through ReLU functions, slightly negative numbers are truncated to zero, while slightly positive ones are left unchanged. These small deviations sometimes compound through a sort of ratcheting effect, in which they accumulate in one of the two networks but not the other. For instance, we observed a value $-2.4980\times10^{-13}$ being truncated to zero by a ReLU, while the corresponding value $2.6903\times10^{-10}$ in the other network remained unchanged as it passed through the ReLU, and repeated operations increased the divergence. Figure~\ref{fig:ptrs_losst_2044} supports this explanation by showing a very slow divergence over training time.

\begin{table*}[h!]
\centering
\begin{tabular}{|c|c|c|c|c|c|c|}
    \cline{2-7}
    \multicolumn{1}{c}{} & \multicolumn{3}{|c|}{PlnNet} & \multicolumn{3}{c|}{ResNet}\\
    \hline
    depth & $\loss$ & $\cost$ & $\error$ & $\loss$ & $ \cost$ & $\error$ \\
    \hline
    20 & 0.04132 & 0.04183 & 0.1121 & 0.04287 & 0.04338 & 0.1131 \\
    \hline
    32 & 0.04880 &  0.04936 & 0.1085 & 0.04847 & 0.04903 & 0.1143 \\
    \hline
    44 & 0.07345 &  0.07407 & 0.1159 & 0.07397 & 0.07459 & 0.1182  \\
    \hline
\end{tabular}
\caption{Training loss $\loss$, training cost $\cost$ and test error $\error$ of equivalently initialized ResNets and PlnNets of depth 20, 32, and 44 after training for 200 epochs on the CIFAR-10 dataset. The initial learning rate was $10^{-2}$, and the rate was divided by 10 after epoch 120 and 160. The parameter $\lambda$ of weight decay is $10^{-4}$. Loss $\loss$ is generally over 50 times more than weight decay term. All the experiments are conducted on the same CPU to guarantee the two networks have exactly the same training process. Figures are reported with four significant decimal digits.}
\label{tab:err_loss_ptrs203244}
\end{table*}

\section{Evolution of Weight Distribution}
\label{dist_ev}
The left column in Figure \ref{fig:hist32} shows two views of the distributions $\chi(z, \bp_0)$ (blue) and $\chi(z, \bt_0)$ (red) of PlnNets and ResNets of depth 32 just after initialization. The plot in the top row shows these two distributions in full with the y-axis clipped to 800. The plot in the bottom row only shows the histogram for the ID entries in the transferred parameter vector $\bt_0$.

The distribution of the ``plain'' weights $\bp_0$ is clearly visible in blue in the overall view at the top. The distribution of the component with zero-mean Gaussian of ``transferred'' weights is also visible in red, but the variance is very small compared to that of the ``plain'' weights. In addition, the distribution of the component with one-mean Gaussian for the transferred weights $\bt_0$ (red) is also visible in the overall view, since the y-axis is clipped to a small number.

After 200 training epochs (right column in Figure \ref{fig:hist32}), the two distributions have not changed much, relative to 1: The blue plots are qualitatively similar to what they were before, and the narrow red peaks, both at 0 and 1, have spread out. However, the second row of the Figure, which displays histograms for the ID entries only, shows that the mode for these entries is still high, around 0.9, in $\bt_{200}$, while very few weights in $\bp$ has magnitude above 0.7 (first row), either before (left: 1 entry bigger than 0.7) or after training (right: 26 entries bigger than 0.7).

Table \ref{tab:weight_mass_p32} shows some additional statistics for the weight distributions, including the total weight mass, defined as the $L_1$ norm of a parameter vector, for (i) all the parameters; (ii) the parameters with magnitude greater than 0.25, and (iii) all ID entries. This table shows in particular that even after training, the total mass of the ID entries in the residual-equivalent parameter vector $\bt_{200}$ still accounts for more than 76 percent ($1000/1300$) of the mass of all the entries greater than 0.25. In contrast, the overall mass of the ID entries in $\bp_0$, initially very small (67), barely changes in $\bp_{200}$, where it increases to 72, a mere 10.29 percent (72/700) of the mass of all the entries greater than 0.25.

\begin{table*}[bt]
\centering
\begin{tabular}{|c|c|*{4}{S[table-format=5.6]|}}
    \cline{3-6}
    \multicolumn{2}{c|}{} & \multicolumn{1}{c|}{$\bp_0$} & \multicolumn{1}{c|}{$\bp_{200}$}
        & \multicolumn{1}{c|}{$\bt_0$} & \multicolumn{1}{c|}{$\bt_{200}$} \\
    \hline
    \multicolumn{2}{|c|}{Mean} & 0.000062 & -0.0061 & 0.0023 & -0.0056 \\
    \hline
    \multicolumn{2}{|c|}{Standard Deviation} & 0.070 & 0.071 & 0.048 & 0.071 \\
    \hline
    \multirow{3}{*}{Mass}& All Weights & 25000 & 25000 & 1900 & 20000 \\
    \cline{2-6}
    & $|\mbox{Weights}| \geq 0.25$ & 420 & 700 & 1100 & 1300 \\
    \cline{2-6}
    & Weights of ID Entries & 67 & 72 & 1100 & 1000 \\
    \hline
\end{tabular}
\caption{Statistics of the parameter vectors $\bp$ and $\bt$ before and after training for a network of depth 32. Figures are reported with two significant decimal digits.}
\label{tab:weight_mass_p32}
\end{table*}

\newcommand{\colWidth}{0.42\columnwidth}
\newcommand{\figWidth}{0.42\columnwidth}
\begin{figure}[h]
\begin{center}
\begin{tabular}{m{0.5em}|m{\colWidth}|m{\colWidth}}
& \multicolumn{1}{c|}{Before Training} & \multicolumn{1}{c}{After Training} \\\hline
\mbox{} & &\\
\begin{sideways}All Weights\end{sideways} &
\includegraphics[width=\figWidth]{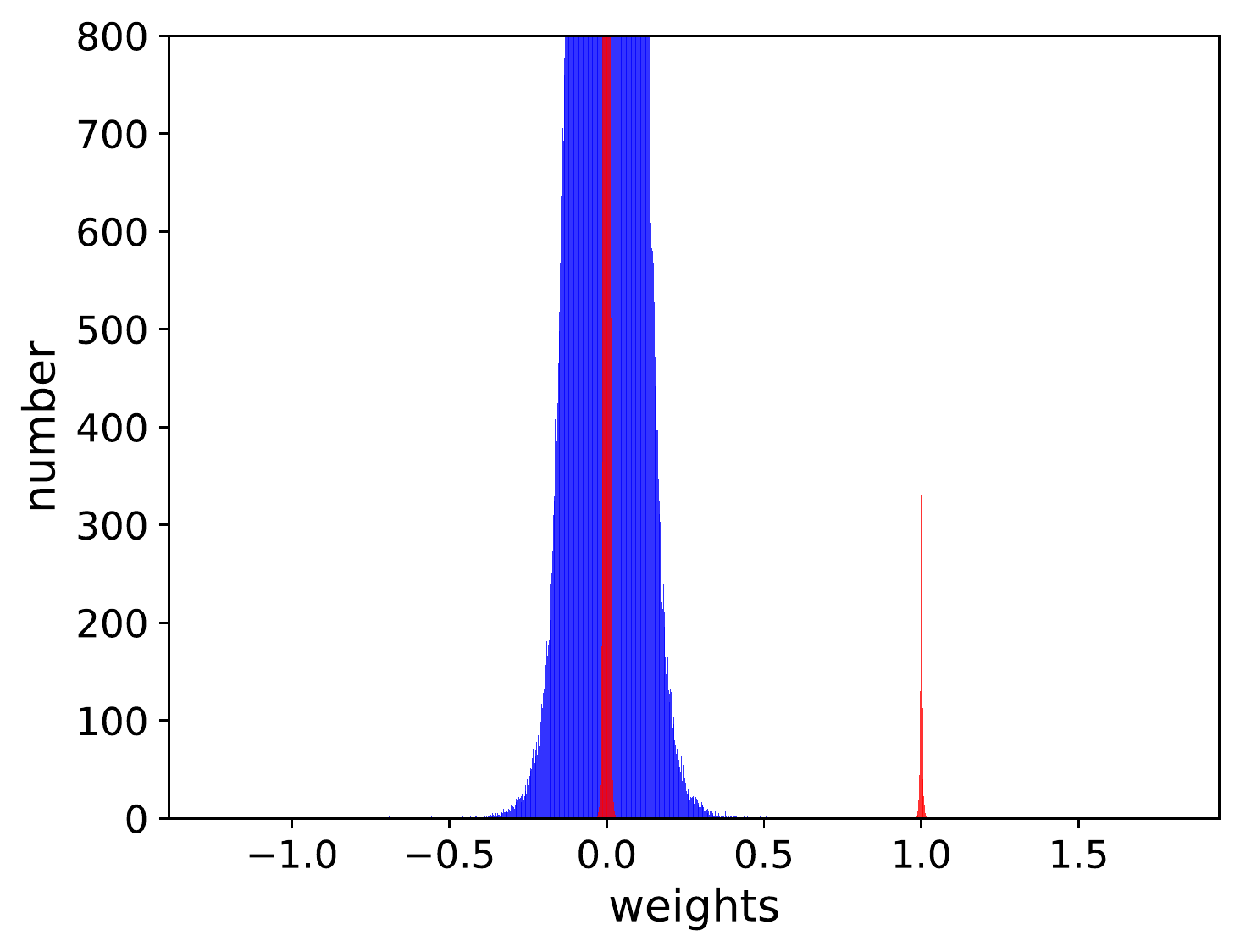} &
\includegraphics[width=\figWidth]{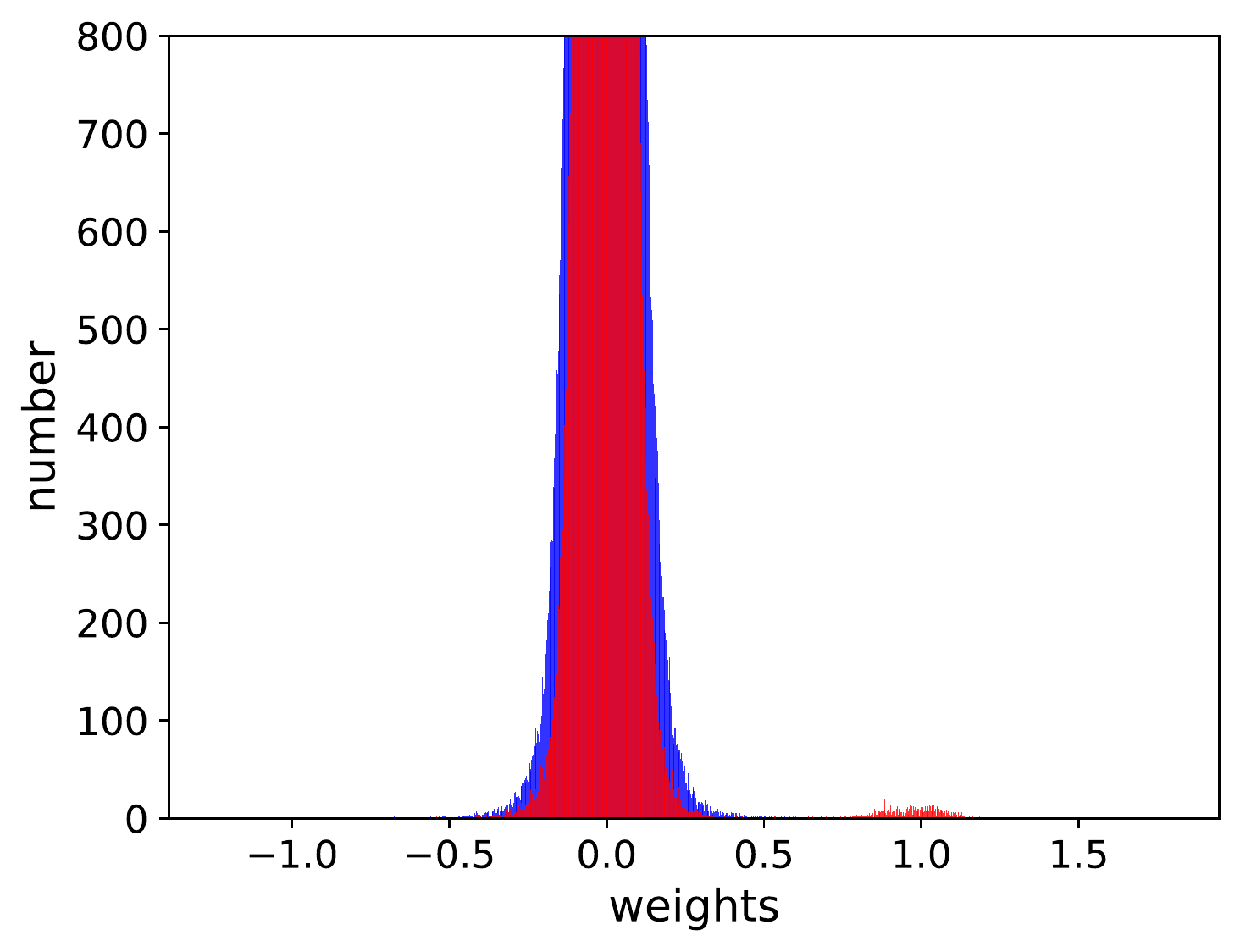} \\\hline
\begin{sideways} Weights of ID Entries \end{sideways} &
\includegraphics[width=\figWidth]{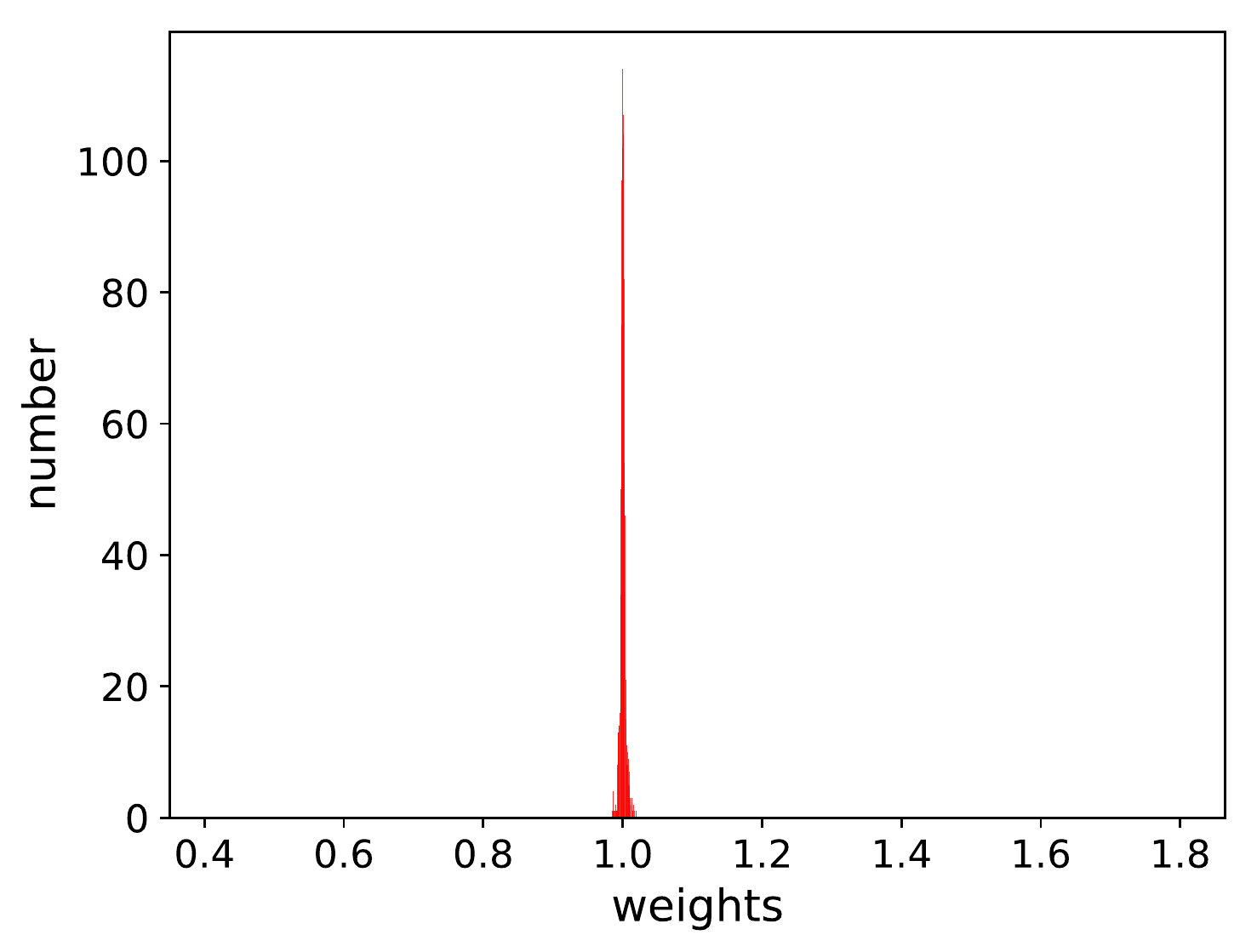} &
\includegraphics[width=\figWidth]{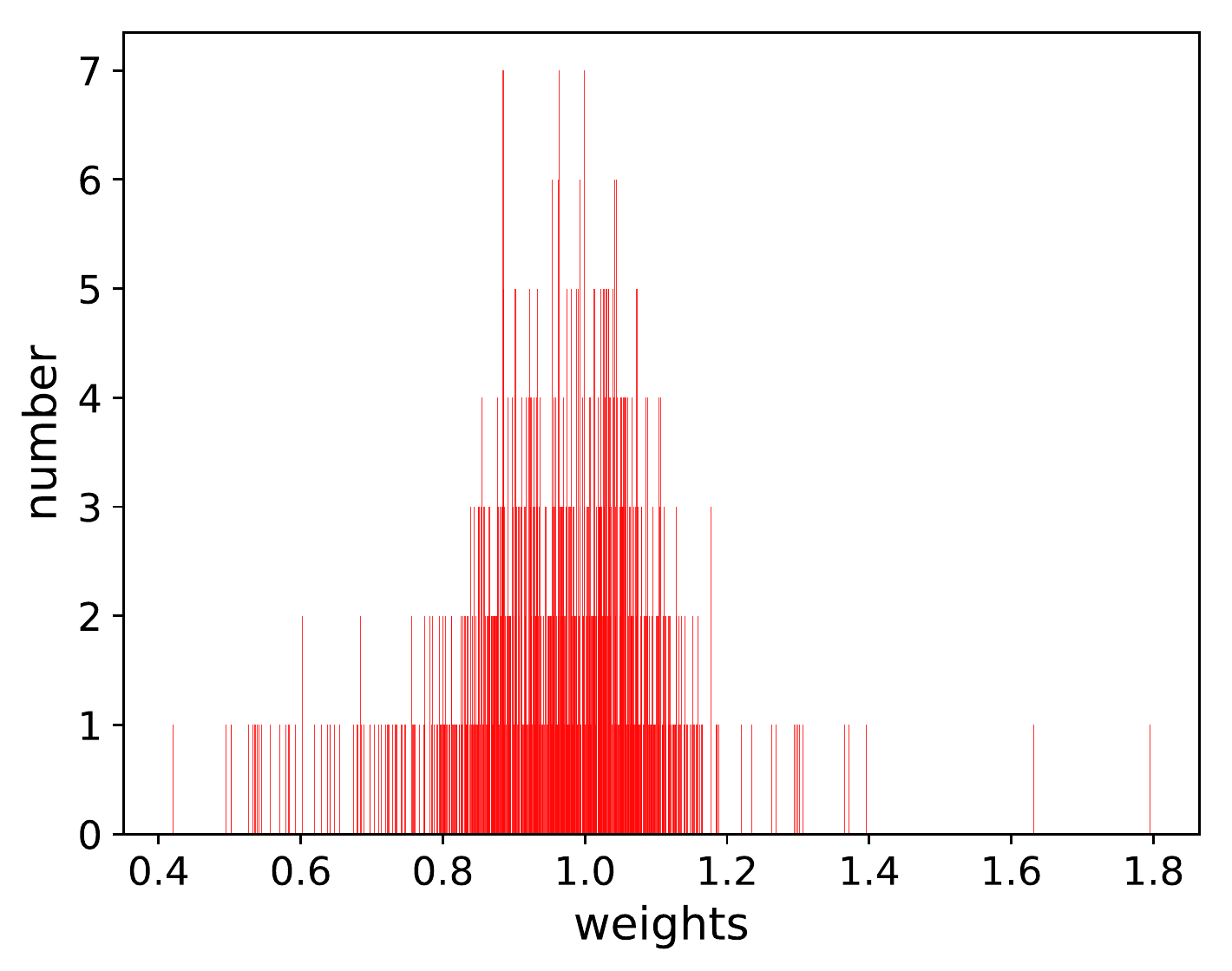} \\\hline
\end{tabular}
\end{center}
\caption{Distribution of all the weights (top row) for depth-32 plain network $p_{32}(\bx, \bp_e)$ (blue) and residual-network-equivalent plain network $p_{32}(\bx, \bt_e)$ (red), and the weights of the ID entries (bottom row) for $p_{32}(\bx, \bt_e)$ before training ($e = 0$, left column) and after 200 epochs ($e = 200$, right column).}
\label{fig:hist32}
\end{figure}

In summary, for this experiment, training modifies the parameters of a network by a small extent, when compared to the differences between the initialization vectors $\bp_0$ and $\bt_0$.

\section{Dominant Gradient Flows}
\label{dom_grad_flow}
In our experiments, residual networks converge to very different and better local minima in training risk than plain networks do after training for 200 epochs, when starting with standard initialization methods.

When the sum $\cost$ (for ``cost'') of risk and weight decay is used as the optimization target, rather than just the risk $\loss$, the weight decay term may play some role in the discrepancy between residual and plain networks. The magnitude of that term is indeed very different (especially initially) for the two types of network. For example, Table~\ref{tab:loss_wr} shows that the weight decay term for a plain network of depth 20 is around 268 ($5.393\times 10^{-4}$/$2.014\times 10^{-6}$) times as large as that of a residual network of the same depth. However, the training loss $\loss$ is in both cases much bigger than the weight decay term, so that the difference in target value introduced by the weight decay term is very small. For example, the initial training loss of a plain network of depth 20 is 4281 (2.309/$5.393\times 10^{-4}$) times as large as the weight decay term, and large ratios occur for plain networks of depths 32 and 44 as well. Therefore, we look for more compelling reasons for the superior performance of residual networks over plain ones.

\begin{table*}[h!]
\centering
\begin{tabular}{|c|c|c|c|c|c|}
    \cline{3-6}
    \multicolumn{2}{c}{} & \multicolumn{2}{|c|}{PlnNet} & \multicolumn{2}{c|}{ResNet}\\
    \cline{2-6}
   	\multicolumn{1}{c|}{} & depth & $\loss$ & weight regularization & $\loss$ & weight regularization \\
    \hline
    \multirow{3}{*}{before training}	& 20 & 2.309 & 0.0005393 & 2.305 & $2.014 \times 10^{-6}$ \\
    \cline{2-6}
    & 32 &  2.308 & 0.0008878 & 2.305 & $2.559 \times 10^{-6}$ \\
    \cline{2-6}
    & 44 &  2.309 & 0.001238 & 2.306 & $3.118\times 10^{-6}$ \\
    \hline
    \multirow{3}{*}{after training}	& 20 & 0.07227 & 0.0007743 & 0.04132 & 0.0005122 \\
    \cline{2-6}
    & 32 & 0.1225 & 0.0009458 & 0.04880 & 0.0005561 \\
    \cline{2-6}
    & 44 & 0.4950 & 0.001180 & 0.07345 & 0.0006194 \\
    \hline
\end{tabular}
\caption{Training loss $\loss$ and weight regularization of PlnNets and ResNets of different depths before training and after training for 200 epochs on the CIFAR-10 dataset. The initial learning rate was $10^{-3}$ for the depth-44 plain network, and $10^{-2}$ for the other 5 networks. The learning rate was divided by 10 after epochs 120 and 160 for all networks. The multiplier $\lambda$ for the weight decay term was $10^{-4}$. The weight decay term is generally very small relative to the training loss $\loss$. The smallest ratio of loss and weight decay, 80.67 (0.04132/$5.122\times 10^{-4}$), is for the ResNet of depth 20 after training.}
\label{tab:loss_wr}
\end{table*}

Two possibilities come to mind. First, there may be no local minimum in terms of training risk in the neighborhood of $\bp_0$ that is as good as the local minimum found in the neighborhood of $\bt_0$. Second, there may exist good local minima in the neighborhood of $\bp_0$, but it may be more difficult, or it may take longer, for plain networks to reach one of them with the same training scheme.

Inspired by two recent articles~\cite{Moritz_Tengyu,kawaguchi2016deep}, we favor the second possibility. These two contributions show that for both types of linear networks all local minima of the risk function are also global minima but there are saddle points in the landscape for linear plain nets. These results indicate that it may be more difficult to train plain networks than residual networks: The plain network may get stuck at a saddle point, and/or may just take longer to reach a minimum. This is consistent with what we observe in our experiments: Figure~\ref{fig:pspt20_losst} (a) shows plots of training risk as a function of training time for a plain network and a residual network, and both are able to achieve a low training risk of $2\times 10^{-3}$ (before full convergence occurs) starting with initial risks around 2.3. However, residual networks almost always get there first.

Starting with this motivation, we propose a conjecture for why residual networks reach a given training risk, say $2\times 10^{-3}$, faster than plain networks do, and why they suffer less from the degradation problem than plain networks do.

\begin{figure}[h]
	\vskip -0.15in
    \centering
    \subfigure[]{\includegraphics[width=0.5\columnwidth]{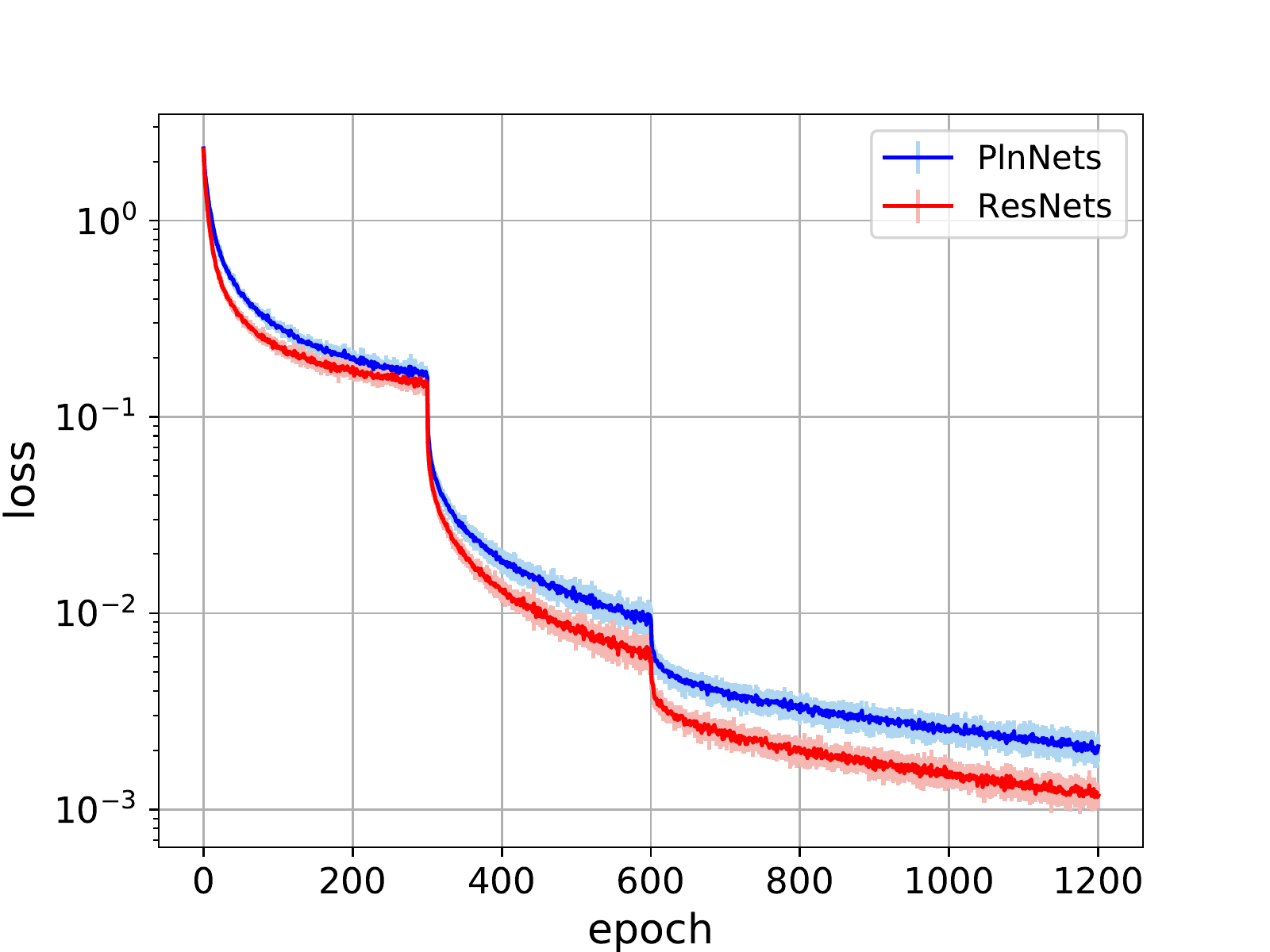}}
    \subfigure[]{\includegraphics[width=0.48\columnwidth]{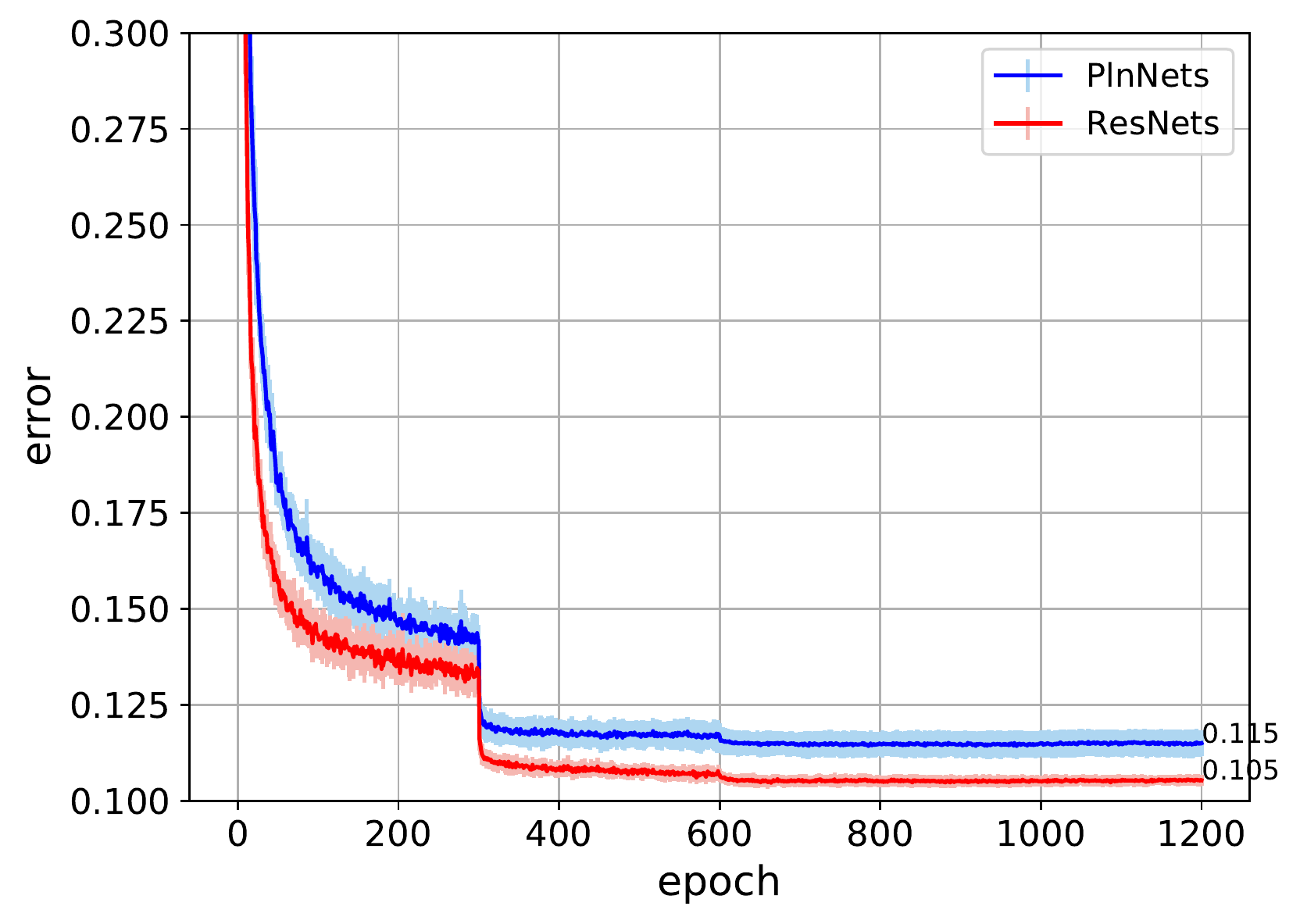}}
    \caption {Plots of risk $\loss$ (left) and error $\error$ (right) as a function of training time for plain (blue) and residual (red) networks of depth 20 after training for 1200 epochs on the CIFAR-10 dataset. The results are based on training 13 plain networks and 13 residual networks initialized at random (but equivalently to each other), to estimate variance in the results. The sequence of training samples are also randomized. Dark color refers to the mean and the half length of the error bar represents the standard deviation.
}
\label{fig:pspt20_losst}
\vskip -0.15in
\end{figure}

\subsection{Simplified Plain Nets Degrade Faster than Simplified Residual Nets}
\label{sec:degradation}

\begin{figure}[h]
\vskip -0.1in
\centering
\subfigure[]{\includegraphics[width=0.48\linewidth]{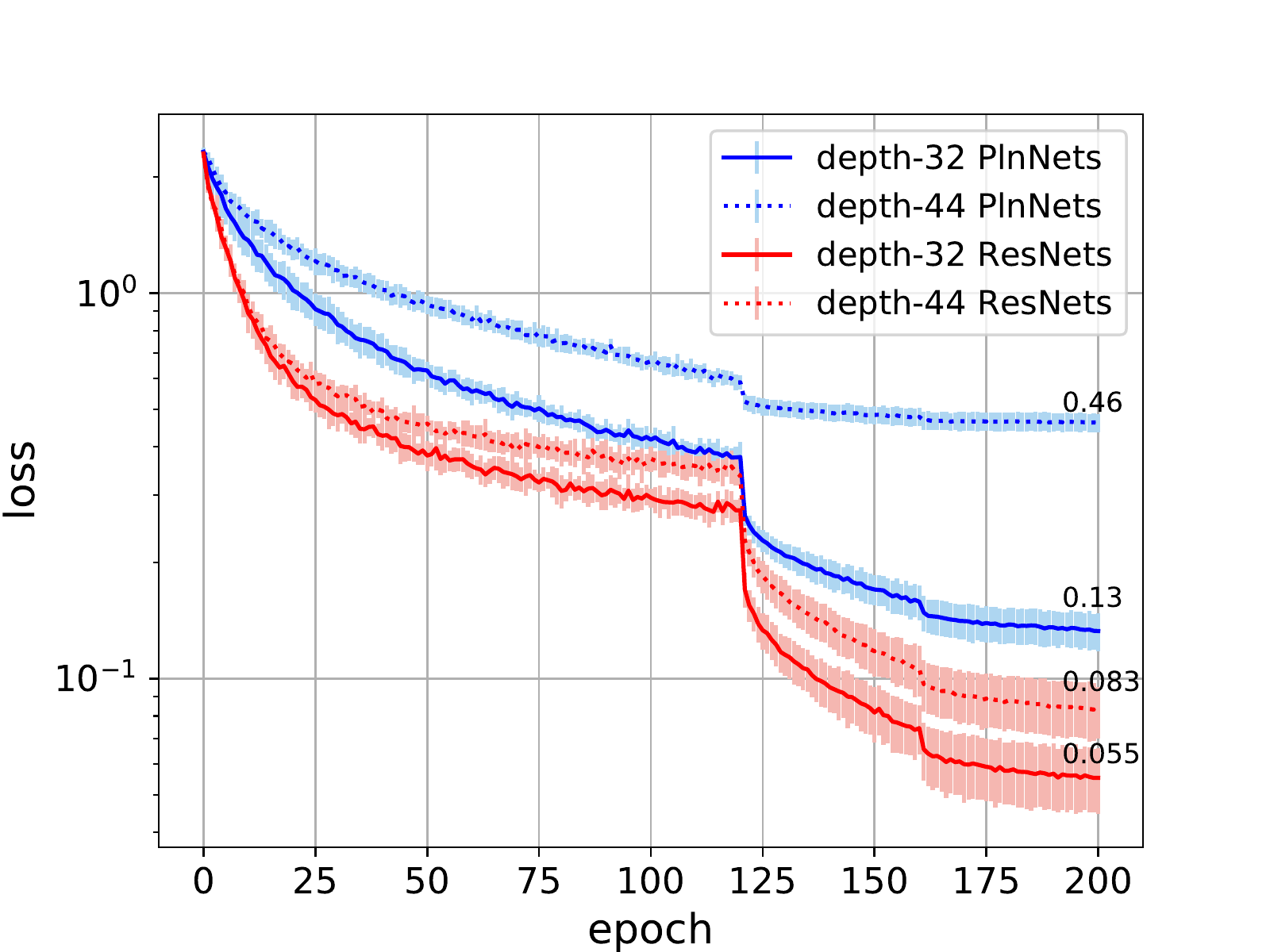}}
\subfigure[]{\includegraphics[width=0.48\columnwidth]{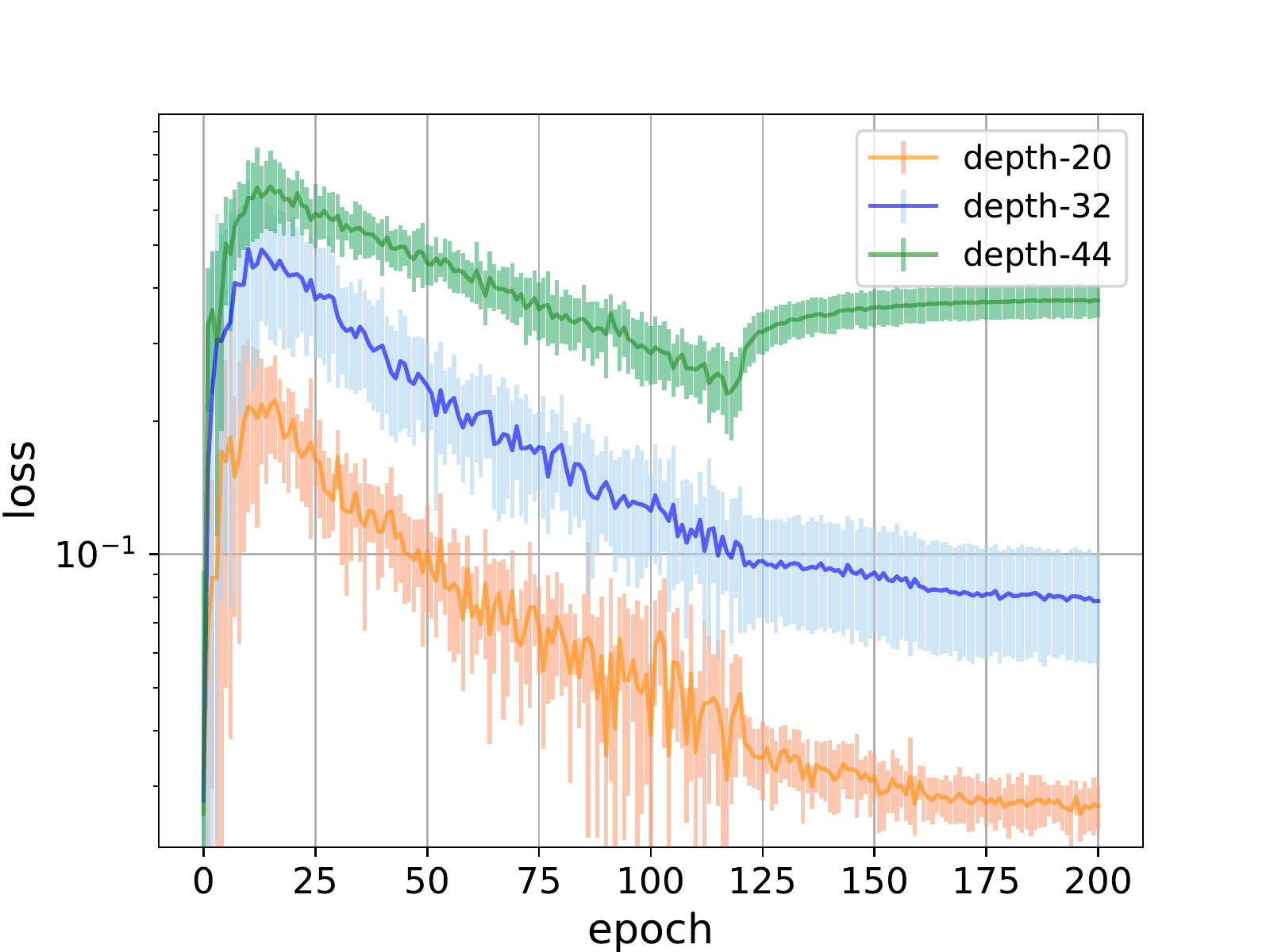}}
\caption {Left: Plots of training risk $\loss$ for plain network $p_{l}(\bx, \bp)$ (blue) and residual-network-equivalent plain network $p_{l}(\bx, \bt)$ (red) of depths $l = $ 32 (solid) and 44 (dotted). The numerical mean after training is attached to each plot. Right: Plots of difference of $\loss$ between plain and residual networks of depth 20, 32, and 44 as a function of training time. For both experiments, the initial learning rate was $10^{-3}$ for depth-44 plain network (larger learning rates do not work) and $10^{-2}$ for the other networks. The y-axis is in logarithmic scale with base 10. The results are based on training 10 randomly initialized networks on random training sample sequences for each network type. Thick lines are means, and the half length of the error bars represents standard deviations.}
\label{fig:pspt_3244_loss_diff_pspt}
\vskip -0.1in
\end{figure}

The diagram in Figure~\ref{fig:pspt_3244_loss_diff_pspt} (a) shows the risk $\loss$ as a function of training epoch for two simplified networks of depth 32 (solid) and 44 (dotted). Each network is in turn trained with two different initialization vectors for 10 times each: The standard KWI vector $\hat{\bp}$ (blue) and the result $\hat{\bt} = \xform(\hat{\br})$ (red) of transferring the standard HMWI initializer $\hat{\br}$ to a plain network. The values of the mean risk after training for 200 epochs are noted explicitly on the diagram.

First, it is clear from the plots that the residual network achieves better training risk than the plain network after a fixed number of epochs, regardless of depth.
Second, the degradation problem is more serious for the plain networks than for the residual networks: Adding 12 layers (from 32 to 44, four new layers in each group $G_{0}$, $G_{1}$, $G_{2}$) makes the risk about 3.54 times worse (0.46/0.13) for plain networks and only about 1.51 times worse (0.083/0.055) for residual networks.

In summary, \emph{simplified networks exhibit the degradation problem as well, and degradation is worse for simplified plain networks than it is for simplified residual networks.}


\subsection{Residual Nets Establish Dominant Signal Paths for Training}
\label{sec:dominance}

Let $\ba^{(\ell)}$ be an \emph{activation vector} that collects all outputs from the ReLU activation function following convolutional layer $\ell$. We subdivide $\ba^{(\ell)}$ into \emph{sub-activations} $\ba_i^{(\ell)}$, one sub-activation vector for each of the $D$ output channels of layer $\ell$. One sub-activation is typically viewed as a two-dimensional feature map, and its entries are listed in some predefined order in $\ba_i^{(\ell)}$.


We define a \emph{Sub-Activation Path (SAP)} as a path connecting sub-activations with the same output channel index $i$. Specifically, the $i^{th}$ sub-activation-path $\SAP_i$ is the sequence $\ba_i^{s(i)}$, $\ba_i^{s(i)+1}$, ..., $\ba_i^{(L)}$, where $s(i) = 1 + gN$ for group $G_g$ and
$N$ is the number of convolutional layers in each group $G_{0}$, $G_{1}$, $G_{2}$. For instance, $N = 10$ for 32-layer networks in our architectures for CIFAR-10 data set. There are 64 final \emph{sub-activations} in $\ba^{(L)}$, which are the outputs of the average layer following the ReLU function, so there are 64 disjoint SAPs. The number of output channels is $D = 16, 32, 64$ in layer groups $G_{0}$, $G_{1}$, $G_{2}$ respectively.

Let now $a_{ip}^{(\ell)}$ be entry $p$ in sub-activation $\ba_i^{(\ell)}$, and recall that each sub-activation $a_j^{(L)}$ in the last convolutional layer is a scalar. Define
\[
\Theta_{ijp}^{(\ell)} = \frac{\partial a_j^{(L)}}{\partial a_{ip}^{(\ell)}} \frac{\partial \loss}{\partial a_j^{(L)}}
\]
to be the component of the derivative of the risk $\loss$ with respect to $\partial a_{ip}^{(\ell)}$ that is mediated by entry $j$ of the final activation $\ba^{(L)}$. We propose the following conjecture.

\begin{quote}
    \emph{The large ID entries of residual networks help create a dominant gradient flow on each of the 64 SAPs, in the sense that for layer $\ell < L$ on SAP $\SAP_i$ we have
    \begin{equation}
    \frac{1}{P} \sum_{p=1}^P\frac{\left| \Theta_{iip}^{(\ell)} \right|}{
    \left| \sum_{j\neq i} \Theta_{ijp}^{(\ell)} \right|} \gg 1
    \label{eq:domCon}
    \end{equation}
    where $P$ is the number of activations in the feature map $\ba_i^{(\ell)}$.
    These dominant gradient flows enhance the convergence speed of residual networks, and do not occur in plain networks.}
\end{quote}

The numerator in each term of the sum in inequality \ref{eq:domCon} measures how much of the supervisory signal flows through sub-activation $i$ in layer $L$ towards sub-activation $a_{ip}^{(\ell)}$, and the denominator measures the magnitude of that signal through the other sub-activations $j\neq i$ in layer $L$. A ratio greater than 1 signals that most of the signal to $a_{ip}^{(\ell)}$ flows through $a_{i}^{(L)}$. The left-hand side of inequality \ref{eq:domCon} is the average ratio, and the conjecture states that in simplified residual networks that ratio is very large, that is, that most of the supervisory signal travels along the mutually disjoint SAPs, rather than ``diagonally'' across other paths in the network. We call the set of network weights along SAP $\SAP_i$ the \emph{effective parameter space} of sub-activation $a_{i}^{(L)}$.



\subsection{Dominant Signal Paths Accelerate Training}
\label{sec:acceleration}

The divergence between solid and dotted blue lines in Figure~\ref{fig:pspt_3244_loss_diff_pspt} (a) shows that adding 12 layers (from 32 to 44) slows down convergence from the very beginning for plain networks, and we attribute this slow-down to the fact that the network's effective parameter space is equal to its entire parameter space. For residual networks, on the other hand, the risk curves (red in Figure~\ref{fig:pspt_3244_loss_diff_pspt} (a)) stay much closer to each other, since the presence of dominant flows restricts the effective parameter spaces to much smaller sets.




Let us construct new kernels for a simplified residual network that discard all the non-dominant flows. There are two types of kernels in our network architecture to consider. The first one is the kernel $T_s$ within each group, which is a square matrix of arrays; the second is the kernel $T_r$ between groups, which is a rectangular matrix (of size $2n\times n$) of arrays, which double the number of channels of output feature maps from the input feature maps. Let us call these new kernels $T' = \{T_s, T_r\}$. For the case of square matrix, we make the kernel $T_s$ a diagonal matrix by making all the non-diagonal entries zero, that is,
\[
T_s = I + R_s\;,
\]
where $R_s$ is a diagonal matrix and the diagonal entries are initialized by HMWI method. The above summation adds one to the central element of each diagonal entry (a 2D array) of $R_s$. For the case of rectangular matrix, both the upper $n\times n$ matrix and lower $n\times n$ matrix of $T_r$ are diagonal matrices that are defined in a similar way as $T_s$. With these new kernels, signals only flow on the dominant signal path. In addition, the number of parameters of the network decreases dramatically. However, such networks still achieve much lower classification error than random guess, which should have $90\%$ error on CIFAR-10 data set. Networks with $T'$ generally achieve more than $76\%$ of the classification accuracy of residual networks of equal depth.

Specifically, Table~\ref{tab:err_loss_TT203244} shows that the classification accuracy of depth-20 networks with $T'$ is around $77.13\% \; ((1-0.3152)/(1-0.1121))$ of that of depth-20 residual network. This indicates that the dominant gradient flows in residual networks are capable of transferring most of the important signals during forward- and back-propagation. In other words, if each sub-activation of $\ba^{(L)}$ is \emph{only} affected by weights in its own effective parameter space, the resulting network can exhibit reasonably good performance after training for the same time as an unrestricted network, say for 200 epochs. The additional weak gradients flows make the final features $\ba^{(L)}$ richer because the additional parameters will affect the final activations even if they make small contributions compared to the dominant gradient flows. If we allow some non-dominant gradient flows, the error will decrease. For example, if the lower $n\times n$ part of the kernel of rectangular shape is not diagonal matrix and each entry is defined by HMWI method, the error of a such depth-20 networks decreases from $31.52\%$ to $21.31\%$.

\begin{table}[h!]
\centering
\begin{tabular}{|c|c|c|c|c|c|c|}
    \cline{2-5}
    \multicolumn{1}{c}{} & \multicolumn{2}{|c|}{Residual Nets} & \multicolumn{2}{|c|}{Network with $T'$}\\
    \hline
    depth & $\loss$ & $\error$ & $\loss$ & $\error$ \\
    \hline
    20 & 0.04132 & 0.1121 & 0.9385 & 0.3152\\
    \hline
    32 & 0.04880  & 0.1085 & 0.9440 & 0.3119\\
    \hline
    44 & 0.07345 & 0.1159 & 0.9397 & 0.3183\\
    \hline
\end{tabular}
\caption{Training loss $\loss$ and test error $\error$ of residual network and network with only dominant gradient flows of depth 20, 32, and 44 after training for 200 epochs on the CIFAR-10 dataset. The initial learning rate was $5\times 10^{-3}$ for depth-44 network with $T'$, and $10^{-2}$ for the other 5 networks. The learning rate was divided by 10 after epoch 120 and 160. The figures are reported with four significant decimal digits.}
\label{tab:err_loss_TT203244}
\end{table}

In addition, there is no obvious degradation problem on the networks with only dominant gradient flows: the loss $\loss$ of depth-44 network is only 1.001 times $(0.9397/0.9385)$ that of the $\loss$ of depth-20 network. This result is consistent with our conjecture, that is, there exist dominant signal paths in residual networks and the signals are propagated mainly on these dominant paths. Since the degradation problem is less severe in residual networks, the increasing number of parameters on networks with $T'$, which only have dominant paths, as the depth of the network increases should not cause serious degradation problem either.

There is a positive correlation between the existence of dominant gradient flow and the difference in training risk between PlnNets and ResNets. Experiments show that the dominant gradient flows are obvious on the whole residual network during the initial stage of training and become weaker as training progresses. The gap in training risk between plain and residual networks widens during this initial period. Specifically, Figure~\ref{fig:pspt_3244_loss_diff_pspt} (b) shows that the difference in training risk between the two networks increases from the beginning of training until around epoch 15. This indicates that the stronger dominant gradient flows enhance the training speed of residual networks.

\subsection{Measurement of Dominant Gradient Flow}
\label{sec:measureFlow}

\subsubsection{Trivial Case}
We first consider the trivial case in which all convolutional kernels are $1\times 1$ and all the activations are 1D vector. To measure dominant gradient flows to layer $l$, we introduce the matrix 
\[
\Theta^{(\ell)} = 
[\prod_{m = l+1}^{L}(K^{(m)})^T S^{(m)}] V
\]
where $S^{(m)} = \frac{\partial \ba^{(m)}}{\partial \bp^{(m)}}$ and $V = diag(\frac{\partial \loss}{\partial \ba^{(L)}})$ are diagonal matrices and $diag(\bx)$ denotes a square diagonal matrix with the elements of vector $\bx$ on its main diagonal. Each diagonal element of $S^{(m)}$ is either zero or one. The $i^{th}$ diagonal element $S^{(m)}_{i, i}$ is zero if $p^{(m)}_{i}$ is negative; otherwise, $S^{(m)}_{i, i}$ is one. Each $(i, j)$ entry $\Theta_{i,j}^{(\ell)}$ measures the amount of gradient of $a_i^{(\ell)}$ that comes from the gradient of $a_j^{(L)}$, that is
\[
\Theta_{i, j}^{(\ell)} = \frac{\partial a_j^{(L)}}{\partial a_i^{(\ell)}} \frac{\partial \loss}{\partial a_j^{(L)}}\;.
\]
Then, we define a vector $\br^{(\ell)}$ of the same size as $\ba^{(\ell)}$ whose $i^{th}$ entry is the ratio
\begin{equation}
    r_{i}^{(\ell)} = \frac{\|\Theta_{i, i}^{(\ell)}\|}
    {\|\sum_{j = 1}^{D^{(L)}}\Theta_{i, j}^{(\ell)} [j \neq i]\|}\;.
    \label{eq:ri}
\end{equation}
The ratio $r_{i}^{(\ell)}$ measures how much of the gradient of $a_i^{(\ell)}$ is from the gradient of $a_i^{(L)}$ when compared to the sum of all other sources. If dominant gradient flows exist from $a_i^{(L)}$ to $a_i^{(\ell)}$, that is if inequality \ref{eq:domCon} holds, then $r_i^{(\ell)}$ is bigger than 1. We define the following measure for the extent to which dominant gradient flows exist from $\ba^{(L)}$ to $\ba^{(\ell)}$:
\[
\sigma^{(\ell)} = \frac{\sum_{i=1}^{|\br^{(\ell)}|}\mathbb{1}[r_i^{(\ell)} > 1]}
{|\br^{(\ell)}|}\;.
\]
Here, $|\br^{(\ell)}|$ is the number of entries in $\br^{(\ell)}$.
We conjecture that the values of $\sigma^{(l)}$ for residual networks are much bigger than those for plain networks for $l = 1, ..., L-1$.

\begin{figure}[h!]
\centering
\includegraphics[width=0.6\columnwidth]{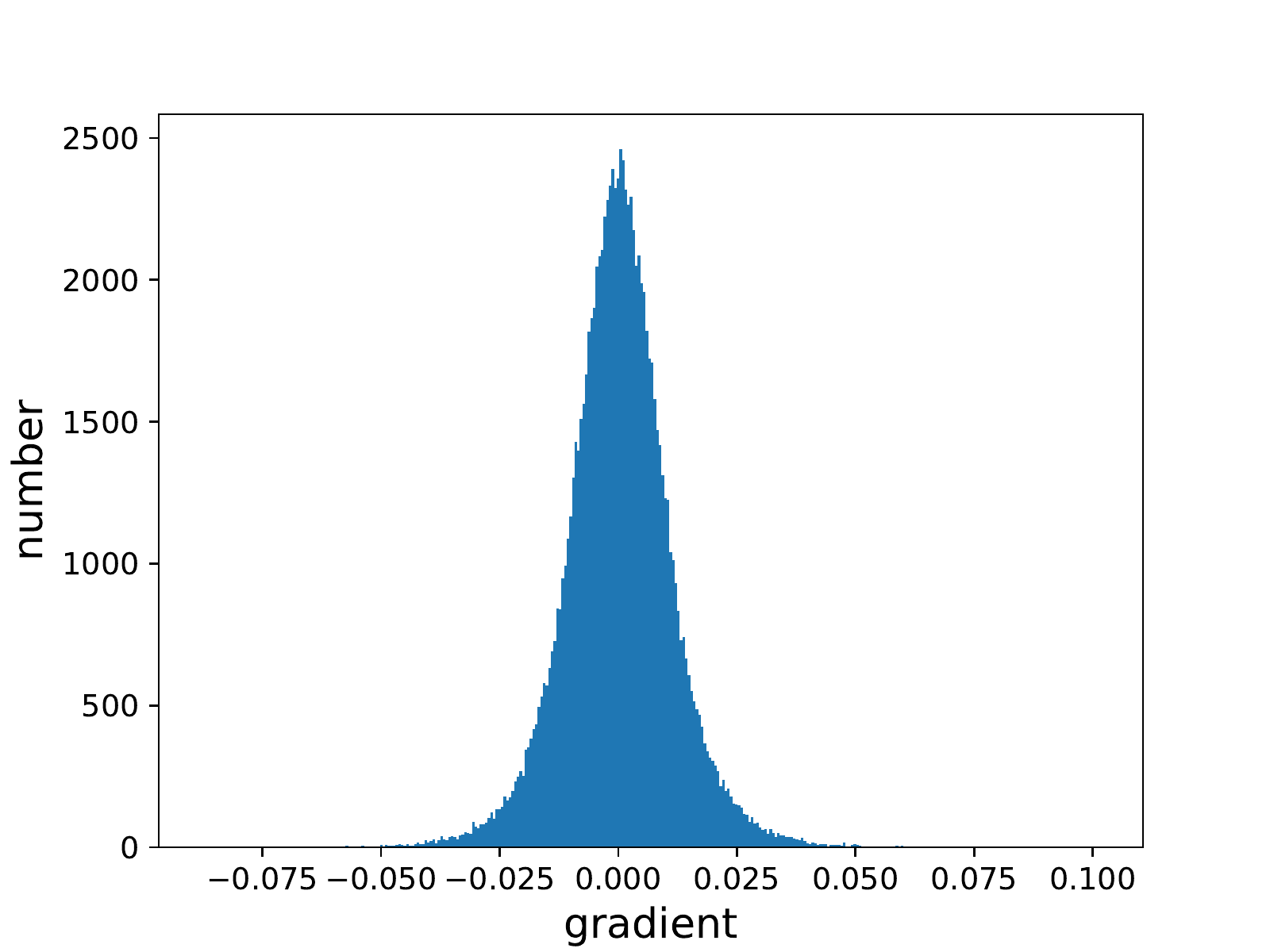}
\caption{Histogram of the gradients of final activation $\ba^{(L)}$ of residual networks. The gradients are collected for 4 epochs (1564 iterations) of training. The mean is $1.44\times 10^{-4}$ and std is $0.011$.}
\label{fig:resnet_finalGradAct_stat}
\end{figure}

\paragraph{Residual Nets:}
We model each weight matrix $K$ as an $n\times n$ square matrix and 
\[
K^{(\ell)} = I + \epsilon A^{(\ell)}\;,
\]
where $\epsilon \ll 1$ and each entry of $A^{(\ell)}$ is an i.i.d. normal random variable with mean 0 and variance $\sigma_A^2$, which is similar to the weights initialization of real networks. Suppose each diagonal of $S^{(\ell)}$ follows a Bernoulli distribution with probability of being 1 as $p^{(\ell)}$ and we fix all $p^{(\ell)}$ to be $p$; suppose each diagonal entry of $V$ is an i.i.d. normal random variable that follows $\mathcal{N}(0, \sigma_V^2)$. Taking the mean of the distribution of entries of $V$ to be zero simplifies the analysis but is still reasonable because the distribution of the final activation of real residual networks follows Normal distribution with mean close to zero as shown in Figure~\ref{fig:resnet_finalGradAct_stat}.

We first consider two extreme cases. The first extreme case is $p = 0$ and thus all the $S$ are zero matrices. Then no gradients can be back-propagated and $\Theta$ is zero matrix. The other extreme case is $p = 1$ and thus all $S$ are identity matrices. Then, by ignoring items that multiply with $\epsilon^m$ for $m > 1$, we have
\[
\Theta^{(\ell)} \approx (I + \epsilon \sum_{m=l+1}^{L} A^{(m)})V = V + \epsilon AV\;,
\]
where $A = \sum_{m=l+1}^{L} A^{(m)}$. Let $B = {\epsilon AV}$; each entry $B_{i,j}$ is the product of two normal random variables. Although the product is no longer normal, we approximate each $B_{i, j}$ as a normal random variable with distribution $\mathcal{N}(0, \epsilon\sqrt{(L-l)}\sigma_A\sigma_V)$ for simplicity.

The denominator of ratio $r_i$ is $R_i = \sum_{j=1}^{n} \Theta_{i,j}\left[j\neq i\right]$, which equals to $\sum_{j=1}^{n}B_{i,j}\left[j\neq i\right]$ in this case; $R_i$ follows normal distribution $\mathcal{N}(0, \sigma_R)$, where $\sigma_R = \epsilon\sqrt{(L-l)(n-1)}\sigma_A\sigma_V$. We have
\[
\mathbb{E}\left[\sigma^{(\ell)}\right] = Pr\left[|\Theta_{i,i}| > |R_i|\right]\;,
\]
where $\Theta_{i,i} = V_{i,i} + B_{i,i}$ and $\Theta_{i,i}$ follows the normal distribution $\mathcal{N}(0, \sqrt{\sigma_V^2+(L-l)\epsilon^2\sigma_A^2\sigma_V^2})$. By Equation~\ref{eq:phi-approx},
\[
Pr\left[|\Theta_{i,i}| > |R_i|\right] \approx \frac{1}{1 + r^{-1.2}}\;,
\]
where 
\[
r = \frac{\sqrt{1+(L-l)\epsilon^2\sigma_A^2}}{\epsilon\sqrt{(L-l)(n-1)}\sigma_A}\;.
\]
This means that $\mathbb{E}\left[\sigma^{(\ell)}\right]$ is not affected by $\sigma_V$.

Between the two extreme cases is $p \in (0, 1)$. By ignoring items that multiply with $\epsilon^m$ for $m > 1$, we have
\begin{equation}
\begin{split}
\Theta^{(\ell)} &= \left[\prod_{m=l+1}^{L}\left(K^{(m)}\right)^T S^{(m)}\right]V \\
&= \left[\prod_{m=l+1}^{L}\left(I + \epsilon A^{(m)}\right)S^{(m)}\right]V \\
&= \left[\prod_{m=l+1}^{L}\left[S^{(m)} + \epsilon A^{(m)}S^{(m)}\right]\right]V \\
&\approx \left(S + \epsilon C\right) V\;,
\end{split}
\end{equation}
where 
\begin{equation}
\begin{split}
S = \left[ \prod_{m=l+1}^{L}S^{(m)} \right], \;\;\;\; C = \sum_{m=l+1}^{L}C^{(m)}, \\
\mbox{and}\;\;\;
C^{(m)} = \left[\prod_{k=l+1}^{m-1}S^{(k)} \right] A^{(m)} \left[\prod_{k=m}^{L}S^{(k)} \right]
\end{split}
\end{equation}

Suppose now that event $Z_i$ means that $S^{(m)}_{ii}$ is 1 for $m = l+1, ..., L$ and $Pr\left[Z_i\right] = p^{L-l}$. Then,
\begin{equation}
\begin{split}
\mathbb{E}\left[\sigma^{(\ell)}\right] &= Pr\left[|\Theta_{i,i}| > |R_i| \right] \\ & \approx Pr\left[|\Theta_{i,i}| > |R_i| \;\Big\vert\; Z_i\right]Pr\left[Z_i\right]\;,
\end{split}
\end{equation}

since it is much less likely that $|\Theta_{i,i}| > |R_i|$ when the complementary event $\bar{Z_i}$ occurs. To calculate $Pr\left[|\Theta_{i,i}| > |R_i| \;\Big\vert\; Z_i\right]$, we need to know random variable $R_i$ and $\Theta_{i,i}$. In this case, $R_i = \epsilon\sum_{j = 1}^{n} D_{i,j}\left[j\neq i\right]$, where $D_{i,j} = C_{i,j}V_{j,j}$. Again, we can approximate $D_{i, j}$ to follow a normal distribution. We know for $j \neq i$,
\begin{equation}
\begin{split}
Pr\left[C^{(m)}_{i,j} = A^{(m)}_{i,j} \;\Big\vert\; Z_i\right] &= p^{L-m+1}\\
Pr\left[C^{(m)}_{i,j} = 0 \;\Big\vert\; Z_i\right] &= 1 - p^{L-m+1}\;.
\end{split}
\end{equation}
Given $Z_i$, in expectation
\[
C_{i,j} = \sum_{m=l+1}^{L} p^{L-m+1}A^{(m)}_{i, j}\;,
\]
Thus, 
\[
\mathbb{Var}[C_{i,j}] = \left[\sum_{m=l+1}^{L} p^{2(L-m+1)}\right]\sigma_A^2 = p'\sigma_A^2\;,
\]
where 
\[
p' = \frac{p^2(1-p^{2(L-l)})}{1-p^2}\;.
\]
Therefore, 
\[
\mathbb{Var}[D_{i,j}] = p'\sigma_A^2\sigma_V^2\;.
\]
Thus, $R_i$ follows the normal distribution $\mathcal{N}(0,\epsilon \sqrt{(n-1)p'}\sigma_A\sigma_V)$. In this case, $\Theta_{i,i} = V_{i,i} + \epsilon D_{i,i}$. We have
\[
Pr\left[C^{(m)}_{i,i} = A^{(m)}_{i,i} \;\Big\vert\; Z_i\right] = 1\;.
\]
Thus, 
\[
\mathbb{Var}[D_{i,i}] = (L-l)\sigma_A^2\sigma_V^2\;,
\]
and therefore 
\[
Pr\left[|\Theta_{i,i}| > |R_i| \;\Big\vert\; Z_i\right] \approx \frac{1}{1 + r^{-1.2}}\;,
\]
where 
\begin{equation}
\begin{split}
r &= \frac{\sigma_V\sqrt{1 + \epsilon^2(L-l)\sigma_A^2}}{\sigma_V\sqrt{\epsilon^2(n-1)p'\sigma_A^2}} = \frac{\sqrt{1 + \epsilon^2(L-l)\sigma_A^2}}{\sqrt{\epsilon^2(n-1)p'\sigma_A^2}} \\
&= \sqrt{\frac{1}{\epsilon^2(n-1)p'\sigma_A^2} + \frac{L-l}{(n-1)p'}}\;.
\end{split}
\end{equation}
Thus,
\[
\mathbb{E}\left[\sigma^{(\ell)}\right] \approx Pr\left[|\Theta_{i,i}| > |R_i| \Big\vert Z_i \right]Pr\left[Z_i\right] \approx \frac{p^{(L-l)}}{1 + r^{-1.2}}\;.
\]
In conclusion,
\[
\mathbb{E}\left[\sigma^{(\ell)}\right] \approx \frac{p^{(L-l)}}{1 + r^{-1.2}}
\]
where
\[
r = \frac{\sqrt{1+(L-l)\epsilon^2\sigma_A^2}}{\epsilon\sqrt{(L-l)(n-1)}\sigma_A}\;.
\]

This result shows directly that the approximate $\mathbb{E}\left(\sigma^{(\ell)}\right)$ increases as $\epsilon$ decreases, $n$ decreases or $\sigma_A$ decreases.

\begin{figure}[h]
    \centering
    \subfigure[]{\includegraphics[width=0.45\columnwidth]{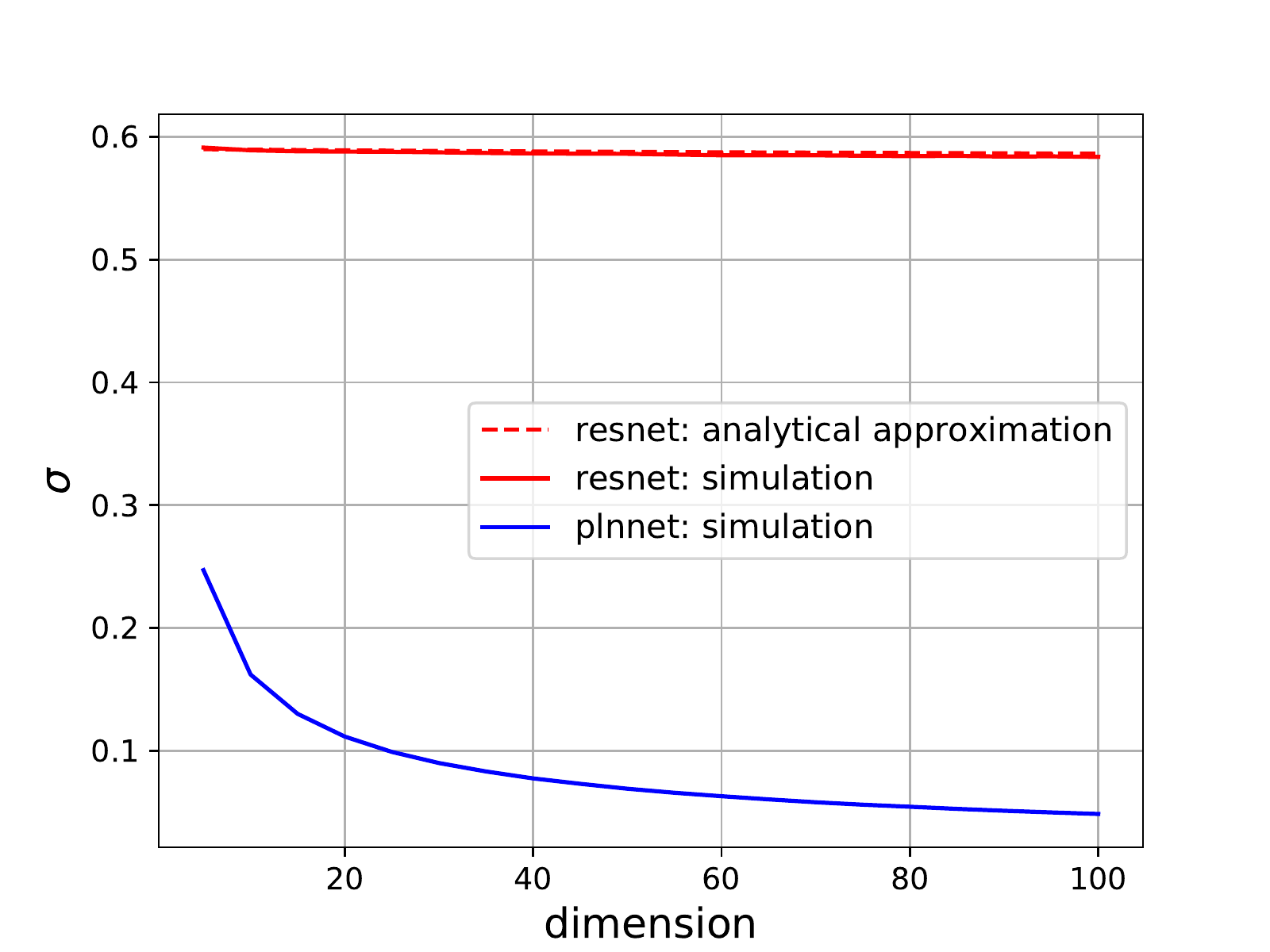}}
    \subfigure[]{\includegraphics[width=0.45\columnwidth]{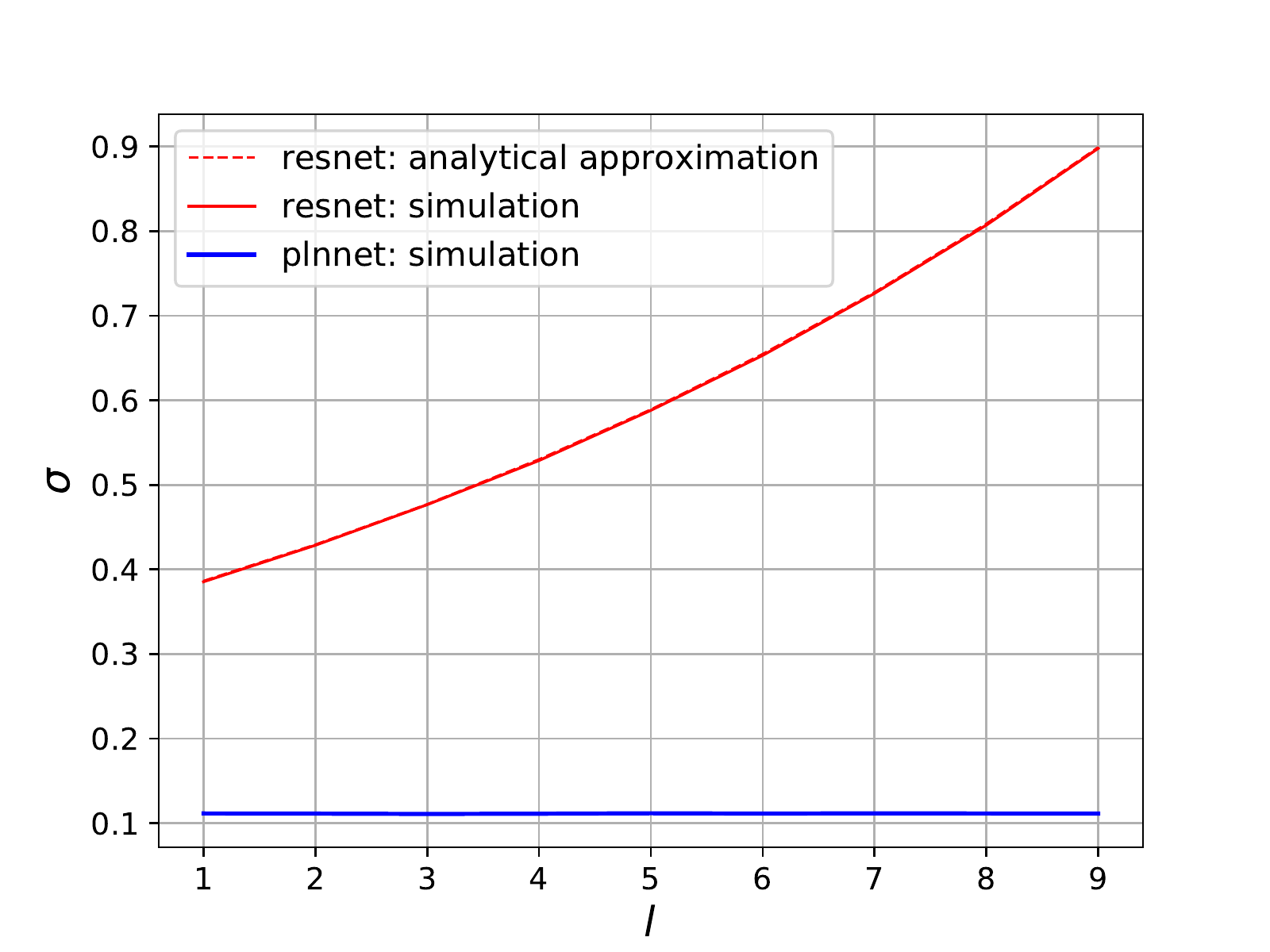}}
    \subfigure[]{\includegraphics[width=0.45\columnwidth]{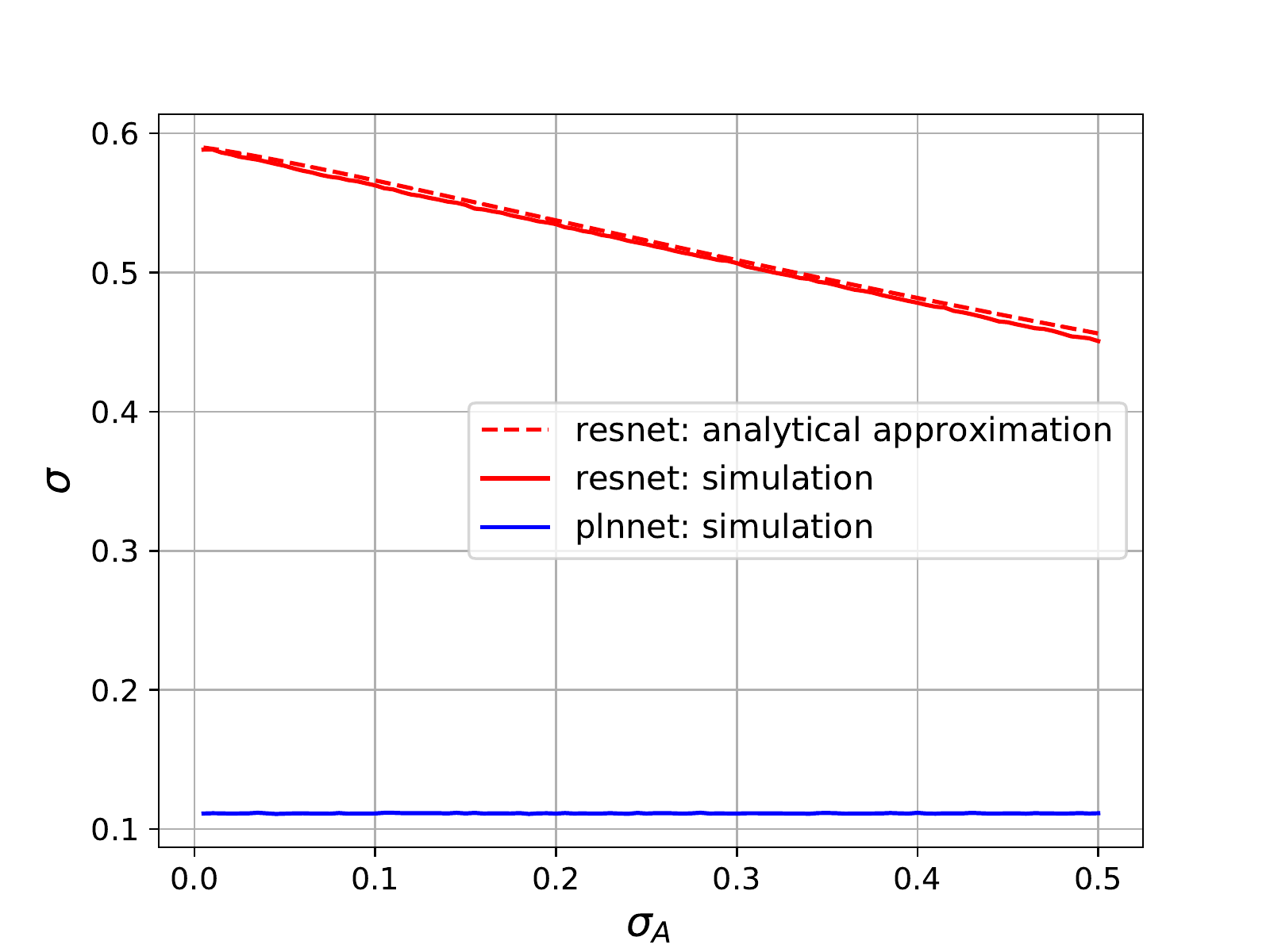}}
    \subfigure[]{\includegraphics[width=0.45\columnwidth]{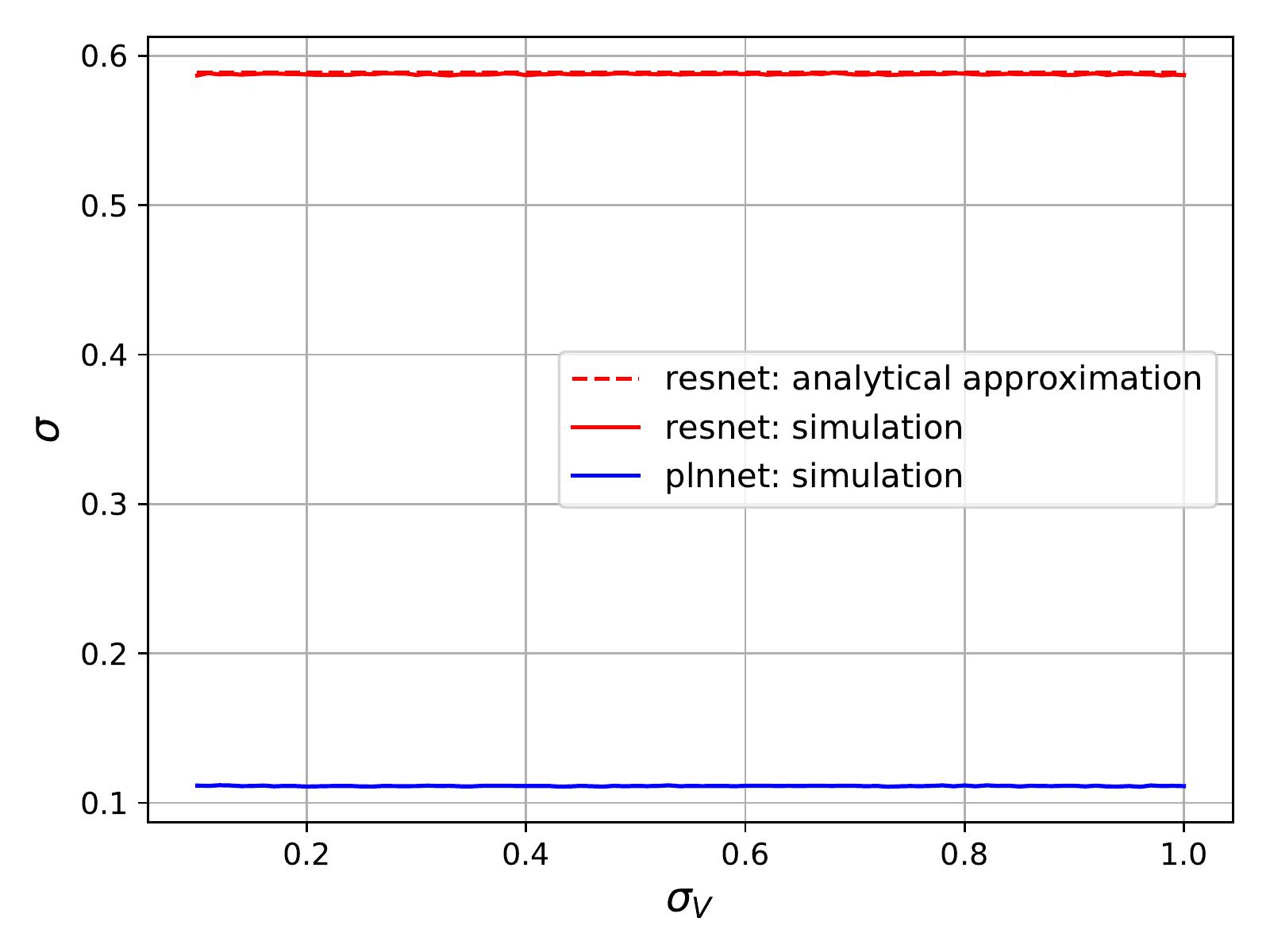}}
    \subfigure[]{\includegraphics[width=0.45\columnwidth]{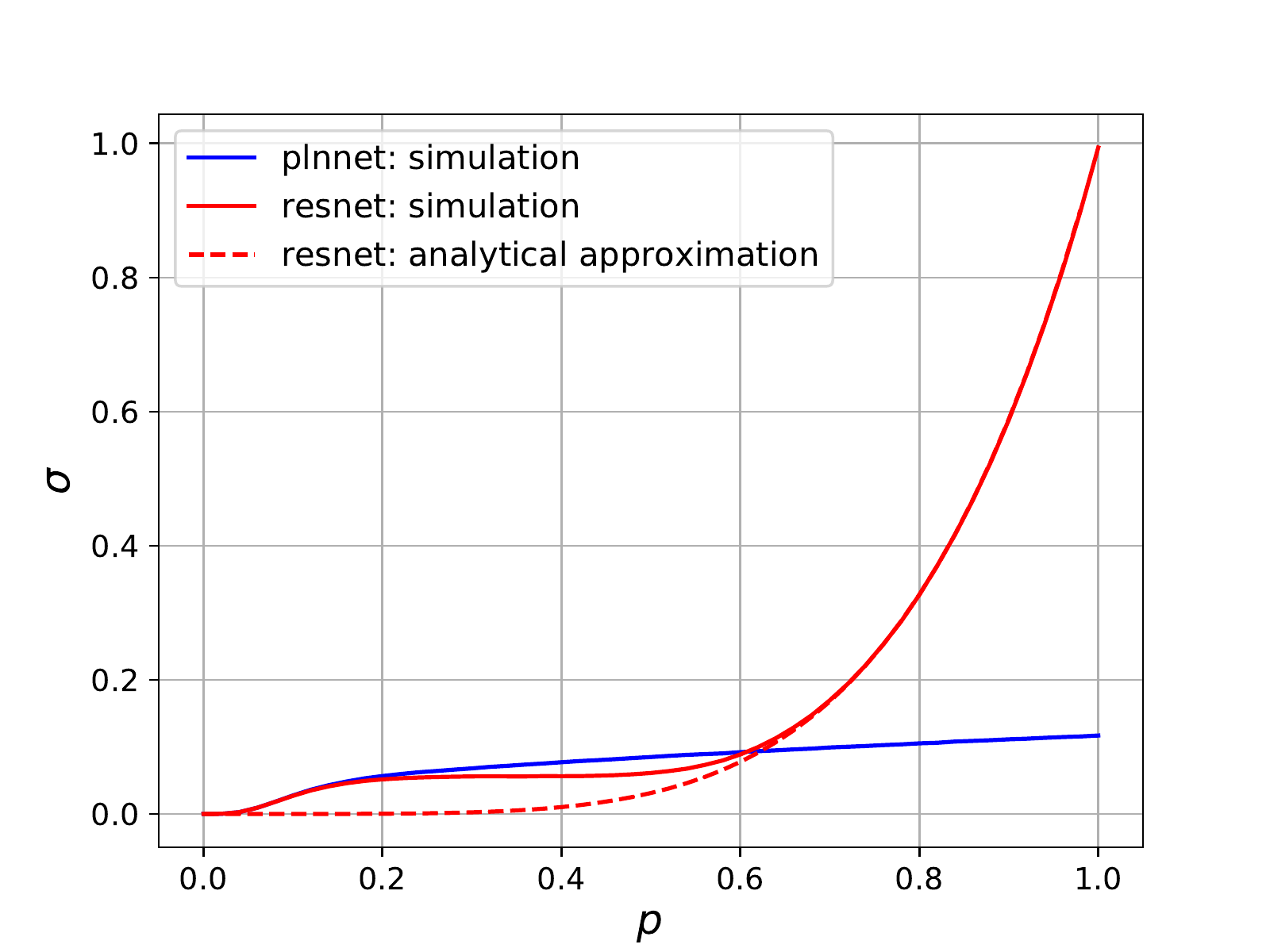}}
    \subfigure[]{\includegraphics[width=0.45\columnwidth]{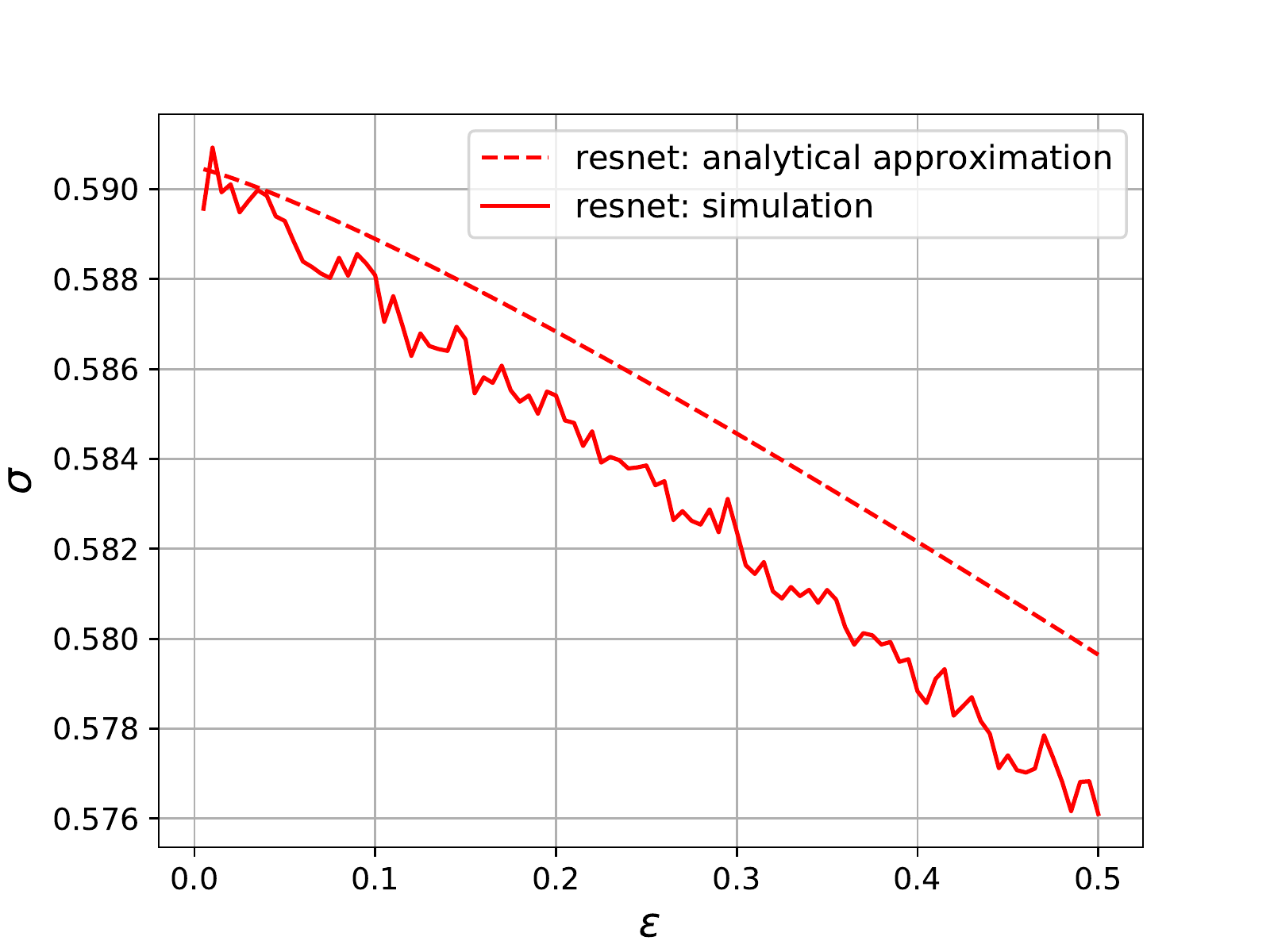}}
  \caption {Plots of expectation of $\sigma$ as a function of (a) dimension $n$ (b) layer index $l$ (c) standard deviation $\sigma_A$ of matrix $A$ (d) standard deviation $\sigma_V$ of the gradients of final activation (e) probability $p$ of being 1 for a diagonal entry of $S$ and (f) scalar $\epsilon$. Blue plots are for plain nets and red plots are for residual nets; red dotted plot represents analytical approximation of the expectation of $\sigma$ of residual nets. Red solid and dotted lines almost coincide in (a) to (d). The expectation of simulated $\sigma$ is estimated by 100K experiments. All the experiments share the same common setting except for the independent variable of each plot: $n = 20$, $\epsilon = 0.1$, total number of layers $L = 10$, $l = 5$, $p=0.9$, $\sigma_A = 0.01$, and $\sigma_V = 0.1$.}
  \label{fig:tri_approx}
  \vskip -0.15in
\end{figure}

Experimental results shown in Figure~\ref{fig:tri_approx} indicate that the approximations made in this analysis are good, as curves for approximate and simulated values for residual nets are very close to each other. In particular, in Figure~\ref{fig:tri_approx} (b) and (e), both the approximate and simulated $\mathbb{E}\left(\sigma^{(\ell)}\right)$ increase as $l$ increases or $p$ increases. In addition, $\sigma_V$ does not have an effect on $\mathbb{E}\left[\sigma^{(\ell)}\right]$, since the red lines are horizontal in Figure~\ref{fig:tri_approx} (d).

\paragraph{Plain Nets:} Each weight matrix $K'$ can be modeled as an $n\times n$ square matrix with each entry following a normal distribution $\mathcal{N}(0, \sigma_A')$. That is,
\[
K' = A'\;.
\]
Similarly, we can define 
\begin{equation}
\begin{split}
\Theta'^{(\ell)} &= 
[\prod_{m = l+1}^{L}(K'^{(m)})^T S'^{(m)}] V' \\
&= [\prod_{m = l+1}^{L}(A'^{(m)})^T S'^{(m)}] V'\;,
\end{split}
\end{equation}
where $S'$ and $V'$ are defined similarly as $S$ and $V$ respectively.
For any $\Theta'^{(\ell)}$, each of its entries follows an identical distribution in expectation because all the matrices are initialised in the same way. In addition, this identical distribution is symmetric around zero. If the distribution is normal, $\sigma^{(\ell)}$ should be only affected by the dimension of $K'$ by Equation~\ref{eq:phi} in the Appendix. However, these variables do not follow a normal distribution. Figure~\ref{fig:tri_approx} (a) to (e) also show the effects of 5 parameters on $\sigma$ of plain nets. The results show that $\sigma^{(\ell)}$ is not only affected by the dimension of $K'$ but also $p$ because the blue line in Figure~\ref{fig:tri_approx} (e) is not horizontal. The cause of $\sigma$ being close to zero of both plain and residual nets when $p$ is small, as shown in Figure~\ref{fig:tri_approx} (e), is that most of the entries in $\Theta$ and $\Theta'$ are zero.

In summary, the existence of dominant gradient flows becomes obvious for residual nets in the trivial case under certain conditions, i.e. $p$, $l$ is big enough and $\epsilon$, $\sigma_A$ is small enough. For example, the dominant gradients flow is not obvious for residual nets when $p < 0.6$ but becomes increasingly obvious as $p$ increases. In addition, the existence of dominant gradient flows is much weaker for plain nets when $p$ is big enough because blue plots are much lower than the red plots in Figure~\ref{fig:tri_approx} (a) to (d). The gap between the trivial case and the networks for CIFAR-10 is that each entry of matrices on the trivial case represents an array in real networks.

\subsubsection{Real Networks}
We can continue to use the following equation to measure the fraction of dominant gradient flows,
\[
\Theta^{(\ell)} = \left[\prod_{m = l+1}^{L}\left(K^{(m)}\right)^T S^{(m)}\right] V\;,
\]
where $S^{(m)} = \frac{\partial \ba^{(m)}}{\partial \bp^{(m)}}$, and $V = diag(\frac{\partial \loss}{\partial \ba^{(L)}})$. $K$, $S$, and $V$ are matrices of arrays in real networks. Matrix $S$ is a diagonal matrix whose diagonal is an array of zeros and ones depending on the sign of the corresponding input to ReLU non-linearity; $V$ is a diagonal matrix of arrays and each diagonal represents the gradients of entries in a channel of the final activation $\ba^{(L)}$. The product of $S^{(m)}_{i, j}$ and $X_{j, k}$ is element-wise multiplication and the product of $K^{(m)}_{i, j}$ and $X_{j, k}$ is 2D convolution operation for any qualified matrix $X$. The resulting $\Theta$ is a matrix of arrays. Suppose $\Theta_{ijp}^{(\ell)}$ is the $p$ entry of the $(i, j)$ entry of $\Theta^{(\ell)}$. Then
\[
\Theta_{ijp}^{(\ell)} = \frac{\partial a_j^{(L)}}{\partial a_{ip}^{(\ell)}} \frac{\partial \loss}{\partial a_j^{(L)}}
\]
where ${a_{ip}^{(\ell)}}$ is the $p$ entry of $\ba_{i}^{(\ell)}$. Then, we define vector $\br^{(\ell)}$ of size $C^l\times 1$ whose $i^{th}$ entry is
\[
r_{i}^{(\ell)} = \frac{1}{P} \sum_{p=1}^P\frac{\left| \Theta_{iip}^{(\ell)} \right|}{\left| \sum_{j\neq i}\Theta_{ijp}^{(\ell)} \right|}
\]

To measure Dominant Gradient Flows in real networks, we use the fraction of existing dominant gradients flows of sub-activation:
\begin{equation}
\sigma^{(\ell)} = \frac{\sum_{i=1}^{\left|\br^{(\ell)}\right|}\mathbb{1}[r_i^{(\ell)} > 1]}
{\left|\br^{(\ell)}\right|}\;
\label{eq:fracGlobalDG}
\end{equation}
where $\mathbb{1}[r_i^{(\ell)} > 1]$ indicates that inequality~\ref{eq:domCon} holds for the $i^{th}$ sub-activation of layer $l$, and $\left|\br^{(\ell)}\right|$ represents the number of sub-activations in that layer. We further average $\sigma^{(\ell)}$ over all training samples in a mini-batch as final result. Our experiments are conducted on CIFAR-10 data set \cite{Krizhevsky2009}. Experimental results show that $\sigma^{(\ell)}$ of ResNets is generally bigger than that of PlnNets through the whole training process for each $l$.

\begin{figure}[h]
    \centering
    \subfigure[]{\includegraphics[width=0.48\columnwidth]{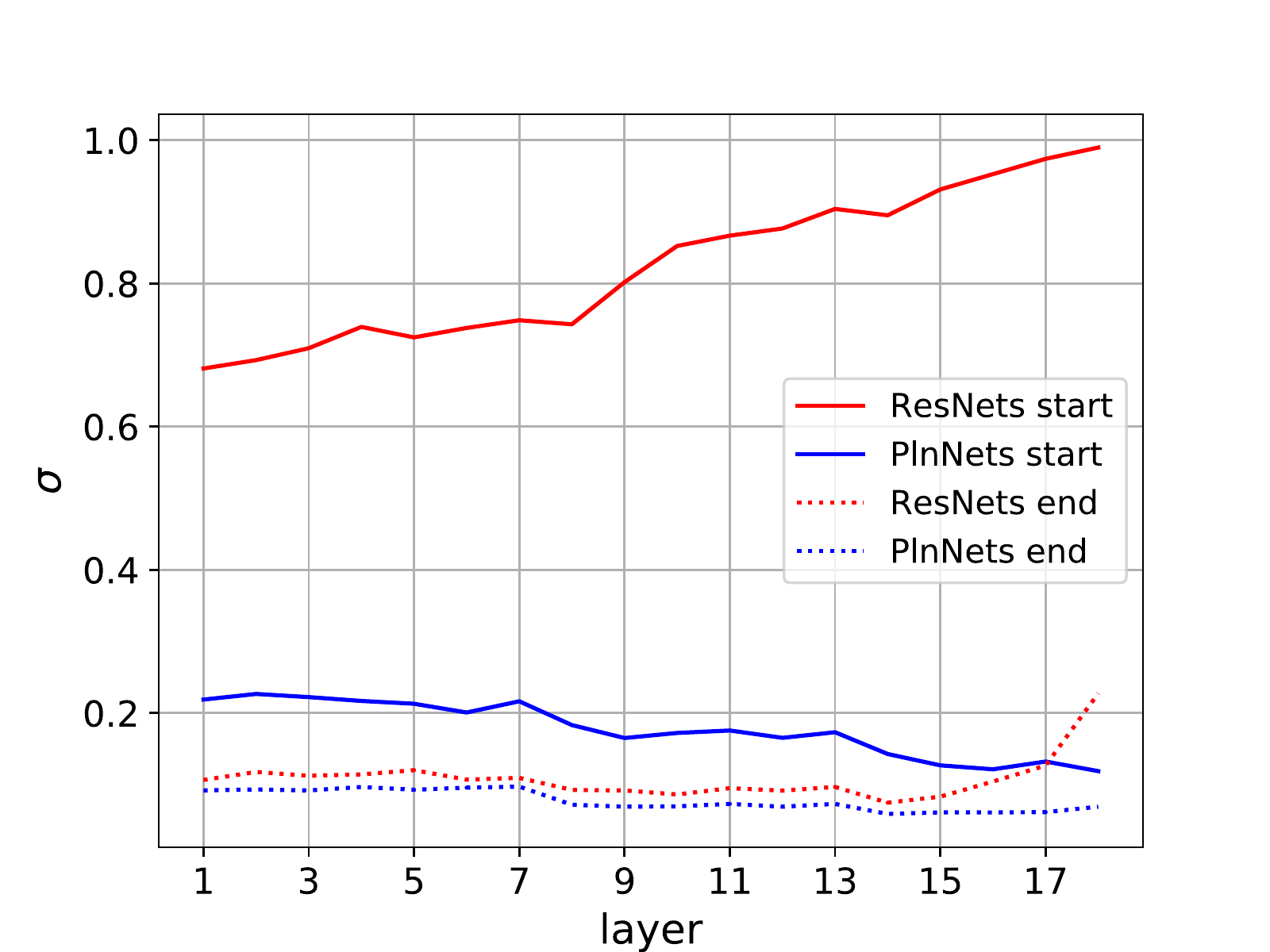}}
    \subfigure[]{\includegraphics[width=0.48\columnwidth]{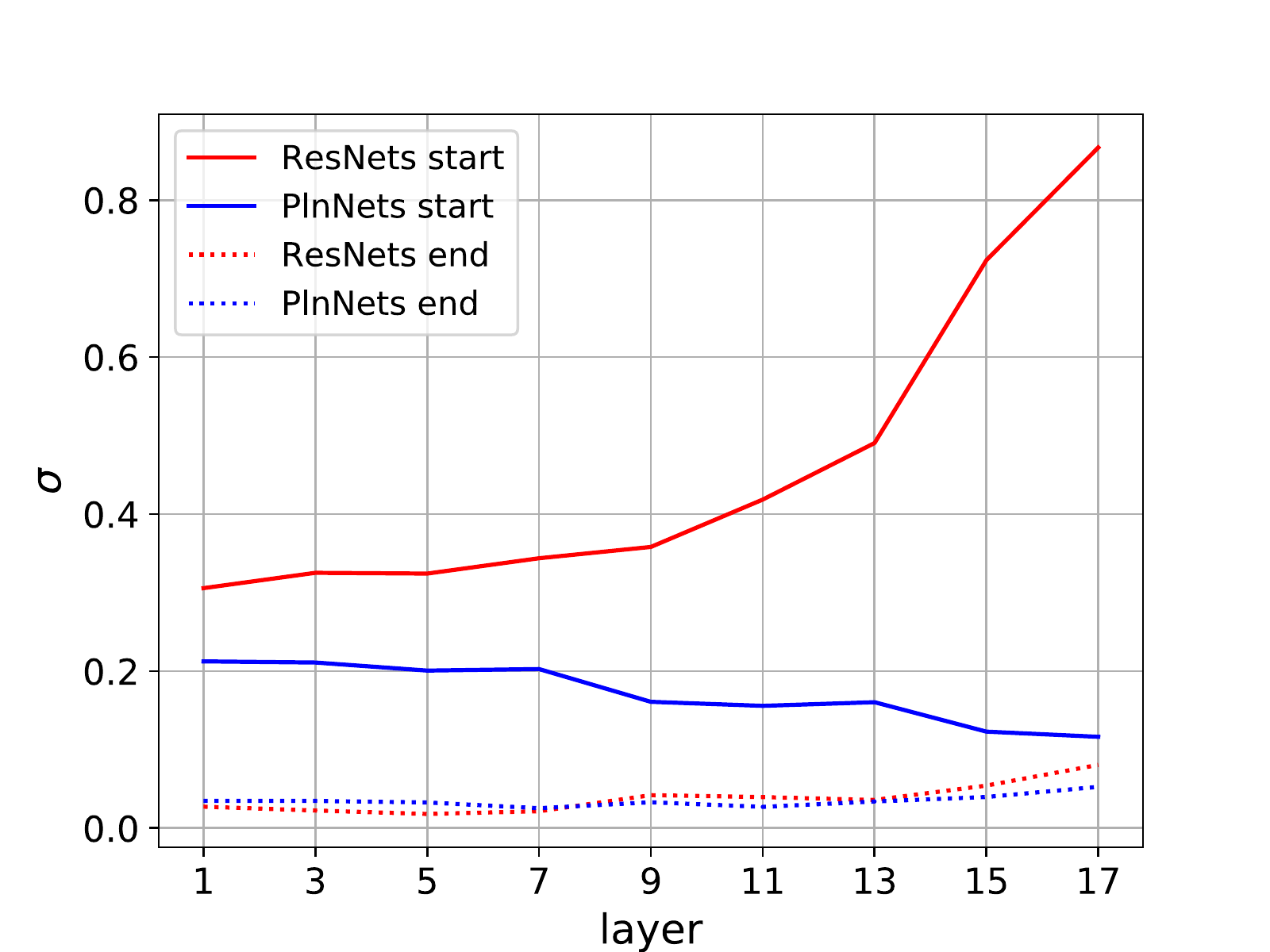}}
  \caption {Plots of fraction $\sigma$ of dominant gradient flows as defined in Equation~\ref{eq:fracGlobalDG} of simplified (left) and fully-fledged (right) depths-20 plain (blue) and residual (red) networks against layers after the first (solid) and last iteration (dotted) of training. The last iteration in diagram (b) refers to the first iteration of the last epoch of training. Simplified and fully-fledged networks are trained for 200 and 180 epochs with initial learning rate 0.01 and 0.1 respectively.}
  \label{fig:DGF_sigma}
\end{figure}

Figure~\ref{fig:DGF_sigma} (a) shows the plots of $\sigma$ on all layers after the first and the last iteration of training for simplified networks. These results show that residual networks (red) have higher $\sigma$ than plain networks (blue) do for all layers. Specifically, $\sigma$ of ResNets is on average 1.24 times of that of PlnNets when $\ell = 1$ and 4.06 times when $\ell = 18$ through the whole training process. We also observe that $\sigma$ increases as $l$ increases for residual networks, which is more obvious on the initial stage of training, e.g. the red solid line increases more obviously as layer increases. This indicates that dominant signal becomes weaker as it flows through more layers. In comparison, the plots for plain networks are flat if not decreasing. This indicates that the gradient of each sub-activation in plain nets is evenly affected by the gradient of each of the final sub-activations regardless of layer.

Another observation is that $\sigma$ of both networks decreases as training continues. In other words, the existence of dominant gradient flows in residual nets is more obvious at the beginning of training, when the risk decreases fastest. This indicates a positive correlation between existence of dominant gradients flow and speed of risk decrease. However, the existence of dominant gradient flows is still strong in deeper layers as training continues: The red dotted line after layer 16 in Figure~\ref{fig:DGF_sigma} (a) is still obviously above the blue dotted line. This also enhances the training of residual networks, although perhaps not to the same extent as during the initial stages.



These results hold also for commonly used fully-fledged residual networks, and not only for the simplified ones. Specifically, Figure~\ref{fig:DGF_sigma} (b) shows the plots of $\sigma$ on all layers after the first and the last iteration of training for fully-fledged networks. We can draw similar conclusions as for simplified networks except for that the dominant gradient flow of residual networks is as weak as that of plain networks in the end of training, and the difference of that between the two networks on shallow layers are smaller. For example, $\sigma$ of ResNets is on average 0.99 times of that of PlnNets through the whole training process for $\ell = 1$. However, the dominant gradient flows in ResNets are still much stronger in deeper layers, e.g. 3.07 times for $\ell = 17$.

\section{Probability of an Inequality between Gaussian Variables}
\label{prob_proof}
\begin{figure}[hbt]
\begin{center}
\includegraphics[width=0.6\columnwidth]{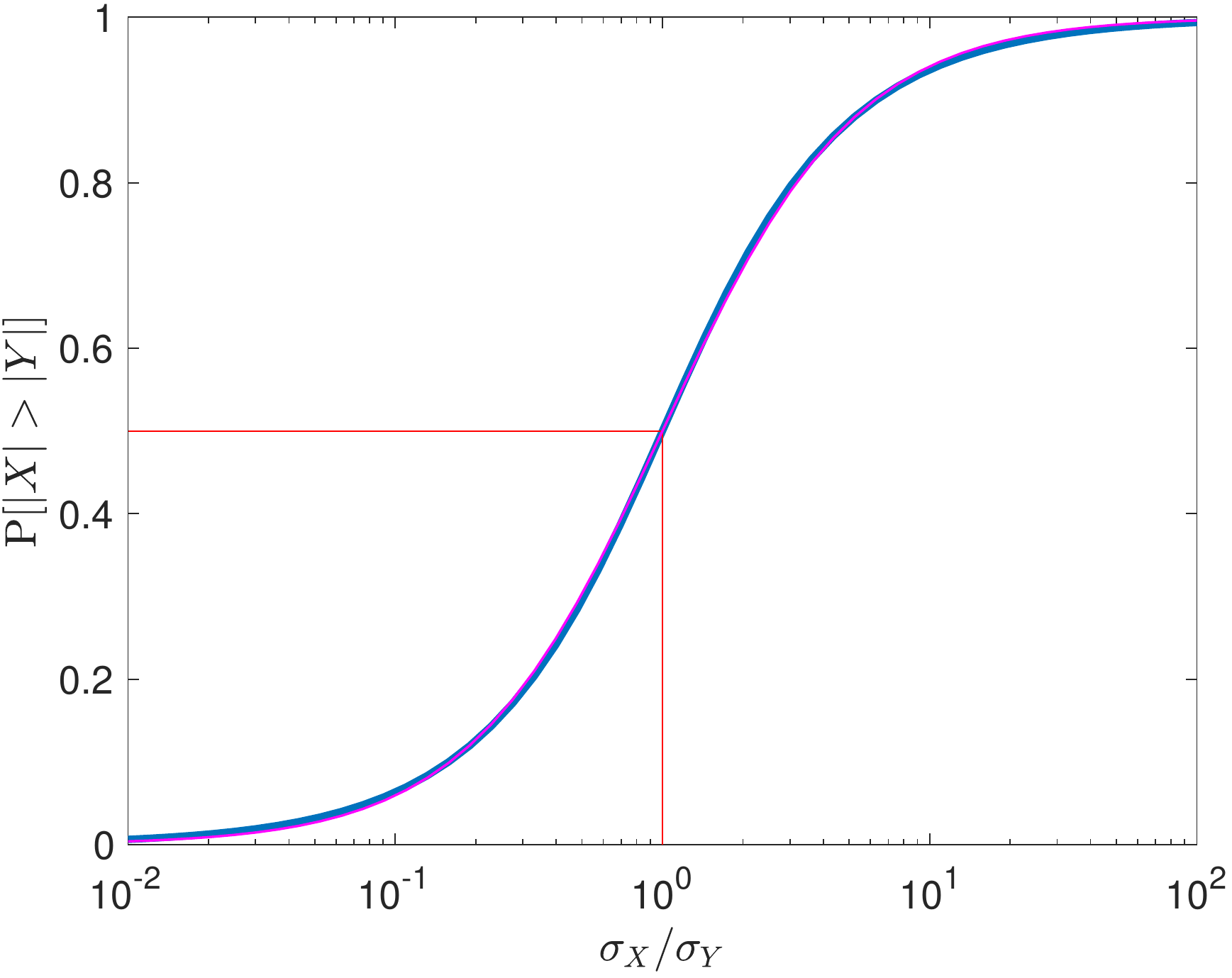}
\end{center}
\caption{Blue plot: The function $\phi(r) = \mathbb{P}[|X| > |Y|]$, where $r = \sigma_X/\sigma_Y$. The two red lines emphasize that $\phi(1) = 1/2$. The (barely visible) thin magenta line is the approximation (\ref{eq:phi-approx}) of $\phi(r)$ for $c = 1.2$.}
\label{fig:phi}
\end{figure}

Let
\[
X \sim \mathcal{N}(0, \sigma_X) \spacetext{and} Y \sim \mathcal{N}(0, \sigma_Y)
\]
be two scalar, independent Gaussian variables with zero mean. Then
\begin{equation}
\mathbb{P}[|X| > |Y|] \;=\; \phi(r) \spacetext{where} r = \frac{\sigma_X}{\sigma_Y}
\label{eq:phi}
\end{equation}
and
\[
\phi(r) \;=\; 4 \int_0^{\infty} g_r(u) G(u) du - 1 \;.
\]
In this expression,
\[
g_{\sigma}(x) = \frac{1}{\sqrt{2\pi}\sigma} e^{-\frac{1}{2} \left(\frac{u}{\sigma}\right)^2}
\]
is the density function of a zero-mean Gaussian variable with standard deviation $\sigma$ and
\[
G(u) = \int_{-\infty}^u g(v) dv
\]
is the cumulative distribution function of the standard Gaussian random variable, with the shorthand
\[
g(v) = g_1(v) \;.
\]
The function $\phi(r)$ is shown in figure \ref{fig:phi}. The fact that the function looks visually similar to a logistic function on a semi-logarithmic plot suggests an approximation of the form
\begin{equation}
\mathbb{P}[|X| > |Y|] \;=\; \phi(r) \approx \frac{1}{1 + r^{-c}} \;.
\label{eq:phi-approx}
\end{equation}
Figure \ref{fig:phi} shows this approximation superimposed as a thin magenta line for $c = 1.2$.

\paragraph{Proof of Equation (\ref{eq:phi}).} Let
\[
p_X(x) = g_{\sigma_X}(x) = \frac{1}{\sigma_X}\ g\left(\frac{x}{\sigma_X} \right)
\]
and
\[
p_Y(y) = g_{\sigma_Y}(y) = \frac{1}{\sigma_Y}\ g\left(\frac{y}{\sigma_Y} \right)
\]
be the probability densities of $X$ and $Y$, and let
\[
p = \mathbb{P}[|X| > |Y|] \;.
\]

 Then, conditioning on the value of $X$ and symmetry of the Gaussian density yields
\begin{eqnarray*}
p &=& \int_{-\infty}^0 p_X(x) \left[ \int_x^{-x} p_Y(y) dy \right] dx  \\
&& \;+\; \int_0^{\infty} p_X(x) \left[ \int_{-x}^x p_Y(y) dy \right] dx \\
&=& 4 \int_0^{\infty} p_X(x) \left[ \int_0^x p_Y(y) dy \right] dx \\
&=& \frac{4}{\sigma_X\sigma_Y} \int_0^{\infty} g\left(\frac{x}{\sigma_X}\right)
\left[ \int_0^x g\left(\frac{y}{\sigma_Y}\right) dy \right] dx \;.
\end{eqnarray*}

Since
\[
\int_0^x g\left(\frac{y}{\sigma_Y}\right) dy \;=\; \sigma_Y \left[G\left(\frac{x}{\sigma_Y} \right) - \frac{1}{2}\right] \;,
\]
and
\[
\int_0^{\infty} g \left(\frac{x}{\sigma_X} \right) dx \;=\; \frac{\sigma_X}{2}\;,
\]
we can write
\begin{eqnarray*}
\frac{\sigma_X \sigma_Y}{4}\ p &=& \sigma_Y \int_0^{\infty} g \left(\frac{x}{\sigma_X}\right)
G \left(\frac{x}{\sigma_Y} \right) dx \\
&& -\frac{\sigma_X \sigma_Y}{4}
\end{eqnarray*}
so that
\[
\frac{p+1}{4} \;=\; \frac{1}{\sigma_X} \int_0^{\infty} g \left(\frac{x}{\sigma_X}\right)
G \left(\frac{x}{\sigma_Y} \right) dx\;.
\]
The change of variable
\[
u \;=\; \frac{x}{\sigma_Y}
\]
and the definition
\[
r \;=\; \frac{\sigma_X}{\sigma_Y}
\]
then yields
\begin{eqnarray*}
\frac{p+1}{4} &=& \frac{\sigma_Y}{\sigma_X} \int_0^{\infty} g \left(\frac{\sigma_Y}{\sigma_X}\, u\right)
G (u)\ du \\
&=& \int_0^{\infty} \frac{1}{r}\ g \left(\frac{u}{r}\right) G (u)\ du
\end{eqnarray*}
that is,
\[
\frac{p+1}{4} \;=\; \int_0^{\infty} g_r(u) G(u) du\;.
\]
Rearranging terms yields the desired result.

\end{document}